\documentclass[10pt,journal,compsoc]{IEEEtran}

\usepackage{graphicx}
\usepackage{amsmath}
\usepackage{amssymb}
\usepackage{booktabs}
\usepackage{microtype}
\usepackage{pifont}
\usepackage{caption}
\usepackage{enumitem}
\usepackage{ragged2e}
\usepackage[table]{xcolor}
\usepackage{subcaption}

\usepackage{bm}
\usepackage{algorithm,algorithmicx,algpseudocode}

\usepackage[pagebackref=false,breaklinks=true,letterpaper=true,colorlinks,bookmarks=false,citecolor=citecolor,linkcolor=linkcolor]{hyperref}

\definecolor{citecolor}{HTML}{0071BC}
\definecolor{linkcolor}{HTML}{ED1C24}

\newcommand{\cmark}{\ding{51}}

\newcommand{\red}[1]{\textcolor{red}{#1}}
\newcommand{\blue}[1]{\textcolor{blue}{#1}}

\ifCLASSOPTIONcompsoc
  \usepackage[nocompress]{cite}
\else
  \usepackage{cite}
\fi

\begin{document}

\title{H\textsubscript{2}OT: Hierarchical Hourglass Tokenizer for Efficient Video Pose Transformers}

\author{
Wenhao Li, Mengyuan Liu$^{*}$, Hong Liu$^{*}$, Pichao Wang, Shijian Lu, and Nicu Sebe
\IEEEcompsocitemizethanks{

\IEEEcompsocthanksitem Wenhao Li is with the \textsuperscript{1}State Key Laboratory of General Artificial Intelligence, Peking University, Shenzhen Graduate School, China, and also with the \textsuperscript{2}College of Computing and Data Science, Nanyang Technological University, Singapore. 
E-mail: wenhao.li@ntu.edu.sg.

\IEEEcompsocthanksitem Mengyuan Liu and Hong Liu are with the \textsuperscript{1}State Key Laboratory of General Artificial Intelligence, Peking University, Shenzhen Graduate School, China. 
E-mail: \{liumengyuan, hongliu\}@pku.edu.cn. 

\IEEEcompsocthanksitem Pichao Wang is with \textsuperscript{3}Amazon AGI, USA. The work does not relate to author's position at Amazon.
E-mail: pichaowang@gmail.com. 

\IEEEcompsocthanksitem Shijian Lu is with the \textsuperscript{2}College of Computing and Data Science, Nanyang Technological University, Singapore. 
E-mail: Shijian.Lu@ntu.edu.sg. 

\IEEEcompsocthanksitem Nicu Sebe is with the \textsuperscript{4}University of Trento, Italy. 
E-mail: niculae.sebe@unitn.it. 

\IEEEcompsocthanksitem $^{*}$Corresponding Authors: Mengyuan Liu, Hong Liu. 
}
}

\markboth{IEEE Transactions on Pattern Analysis and Machine Intelligence}
{Li \MakeLowercase{\textit{et al.}}: 
\title{H\textsubscript{2}OT: Hierarchical Hourglass Tokenizer for Efficient Video Pose Transformers}
}

\IEEEtitleabstractindextext{
\justify
\begin{abstract}
Transformers have been successfully applied in the field of video-based 3D human pose estimation. However, the high computational costs of these video pose transformers (VPTs) make them impractical on resource-constrained devices. In this paper, we present a hierarchical plug-and-play pruning-and-recovering framework, called \textbf{H}ierarchical \textbf{Ho}urglass \textbf{T}okenizer (H\textsubscript{2}OT), for efficient transformer-based 3D human pose estimation from videos. H\textsubscript{2}OT begins with progressively pruning pose tokens of redundant frames and ends with recovering full-length sequences, resulting in a few pose tokens in the intermediate transformer blocks and thus improving the model efficiency. It works with two key modules, namely, a Token Pruning Module (TPM) and a Token Recovering Module (TRM). TPM dynamically selects a few representative tokens to eliminate the redundancy of video frames, while TRM restores the detailed spatio-temporal information based on the selected tokens, thereby expanding the network output to the original full-length temporal resolution for fast inference. Our method is general-purpose: it can be easily incorporated into common VPT models on both \textit{seq2seq} and \textit{seq2frame} pipelines while effectively accommodating different token pruning and recovery strategies. In addition, our H\textsubscript{2}OT reveals that maintaining the full pose sequence is unnecessary, and a few pose tokens of representative frames can achieve both high efficiency and estimation accuracy. Extensive experiments on multiple benchmark datasets demonstrate both the effectiveness and efficiency of the proposed method. Code and models are available at \url{https://github.com/NationalGAILab/HoT}. 
\end{abstract}

\begin{IEEEkeywords}
3D Human Pose Estimation, Video Pose Transformer, Token Pruning, Token Recovering
\end{IEEEkeywords}}

\maketitle

\IEEEdisplaynontitleabstractindextext

\IEEEpeerreviewmaketitle

\IEEEraisesectionheading{\section{Introduction}}
\IEEEPARstart{3D} human pose estimation (HPE) from videos has numerous applications, such as action recognition \cite{liu2017enhanced,zhou2024blockgcn,luvizon2020multi}, human-robot interaction \cite{zimmermann20183d,garcia2019human,li2024human}, and computer animation \cite{mehta2017vnect,tu2024motioneditor}. 
Current video-based 3D HPE methods mainly follow the pipeline of 2D-to-3D pose lifting \cite{you2023co,chen2023hdformer,liu2023posynda,zhang2024app}. 
This two-stage pipeline first utilizes an off-the-shelf 2D HPE model to detect 2D body joints for each video frame and then employs a separate lifting model to estimate 3D pose sequences from the detected 2D poses. 

Recently, transformer-based architectures \cite{poseformer,mhformer,mixste,motionbert,liu2025tcpformer} have shown state-of-the-art (SOTA) performance in the field of video-based 3D HPE, thanks to their competency in modeling the long-range dependencies among video frames. 
These video pose transformers (VPTs) typically regard each video frame as a pose token and utilize extremely long video sequences to achieve superior HPE performance (\textit{e.g.}, 81 frames in \cite{poseformer}, 243 frames in \cite{mixste,pstmo,motionbert}, or 351 frames in \cite{stride,mhformer,einfalt2023uplift}). 
However, these methods inevitably suffer from high computational costs since the complexity of the self-attention in VPT grows quadratically with respect to the number of tokens (\textit{i.e.}, frames), hindering the deployment of these heavy VPTs in many real-world applications. 

\begin{figure*}[t]
\centering
\includegraphics[width=0.8\linewidth]{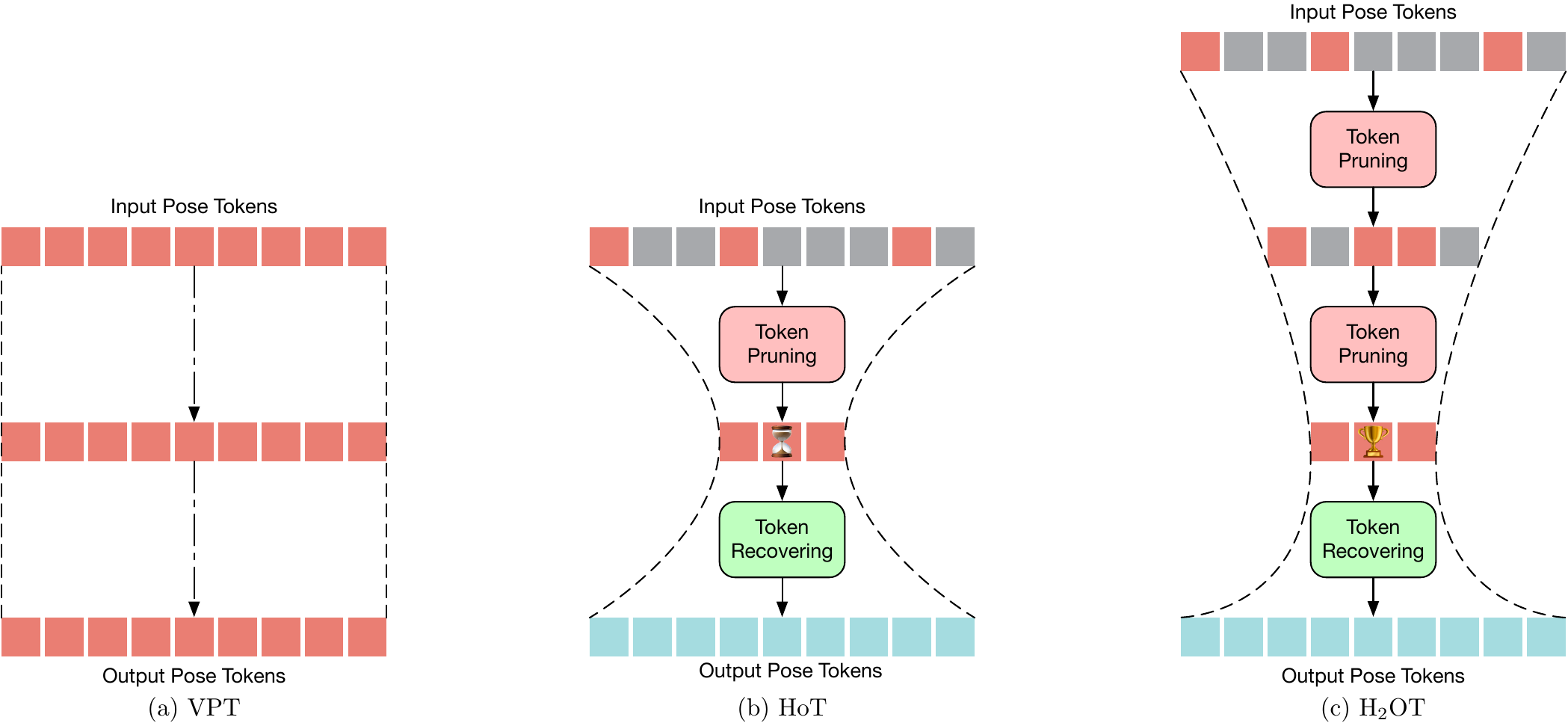}
\caption
{
Illustration of different VTP frameworks. 
\textbf{(a)} Existing VPTs \cite{poseformer,mhformer,mixste,motionbert} follow a ``rectangle'' paradigm that retains the full-length sequence across all blocks, which incurs expensive and redundant computational costs. 
\textbf{(b)} HoT \cite{hot} follows an ``hourglass'' paradigm that prunes the pose tokens and recovers the full-length tokens, which keeps a few tokens in the intermediate transformer blocks and thus improves the model efficiency. 
\textbf{(c)} Our H\textsubscript{2}OT extends HoT \cite{hot} by introducing a hierarchical pruning design, forming an ``trophy-shaped (pyramidal)'' paradigm. 
The gray squares represent the pruned tokens. 
}
\label{fig:pipeline}
\end{figure*}

Two factors are crucial for achieving efficient VPTs. \textbf{First}, directly reducing the frame number can boost VPTs' efficiency, but it results in a small temporal receptive field that hinders the model from capturing sufficient spatio-temporal information in pose estimation \cite{videopose,liu2020attention}. 
Hence, it is essential to design an efficient solution while maintaining a large temporal receptive field for accurate estimation. 
\textbf{Second}, adjacent frames in a video sequence contain redundant information due to the similarity of nearby poses (50 Hz cameras used in Human3.6M \cite{ionescu2013human3}). 
Moreover, recent studies \cite{rao2021dynamicvit,wang2022vtc} found that many tokens tend to be similar in the deep transformer blocks. 
Hence, using full-length pose tokens in these blocks tends to introduce redundant calculations but contributes little to the pose estimation. 

Based on these observations, we propose to prune pose tokens in the deep transformer blocks to improve the efficiency of VPTs. 
Although token pruning can reduce the number of tokens and improve efficiency, it also makes it difficult to estimate the consecutive 3D pose of all frames, as each token corresponds to one frame in existing VPTs \cite{mhformer,mixste,motionbert}. 
Additionally, for efficient inference, a real-world 3D HPE system should be able to estimate the 3D poses of all frames at once in an input video. 
Therefore, it is necessary to recover the full-length tokens to estimate 3D poses for all frames so that the model can achieve fast inference and be compatible with existing VPT frameworks. 

Driven by this analysis, we present a novel hierarchical \textit{pruning-and-recovering} framework for efficient transformer-based 3D HPE from videos. 
Different from existing VPTs \cite{poseformer,mhformer,mixste,motionbert} that maintain the full-length sequence across all blocks, our method begins with progressively pruning the pose tokens of redundant frames and ends with recovering the full-length tokens. 
With such pruning and recovering designs, we can keep only a few tokens in the intermediate transformer blocks and effectively improve the model efficiency (see Fig.~\ref{fig:pipeline}). 
Specifically, we design a Token Pruning Module (TPM) to select a few representative tokens, which can maintain rich information while reducing video redundancy. 
In addition, we design a Token Recovering Module (TRM) to expand the low temporal resolution caused by the pruning operation to the full temporal resolution. 
This strategy enables the network to estimate consecutive 3D poses of all frames and clearly improves the inference speed. 

\begin{figure*}[tb]
\centering
\includegraphics[width=1.00\linewidth]{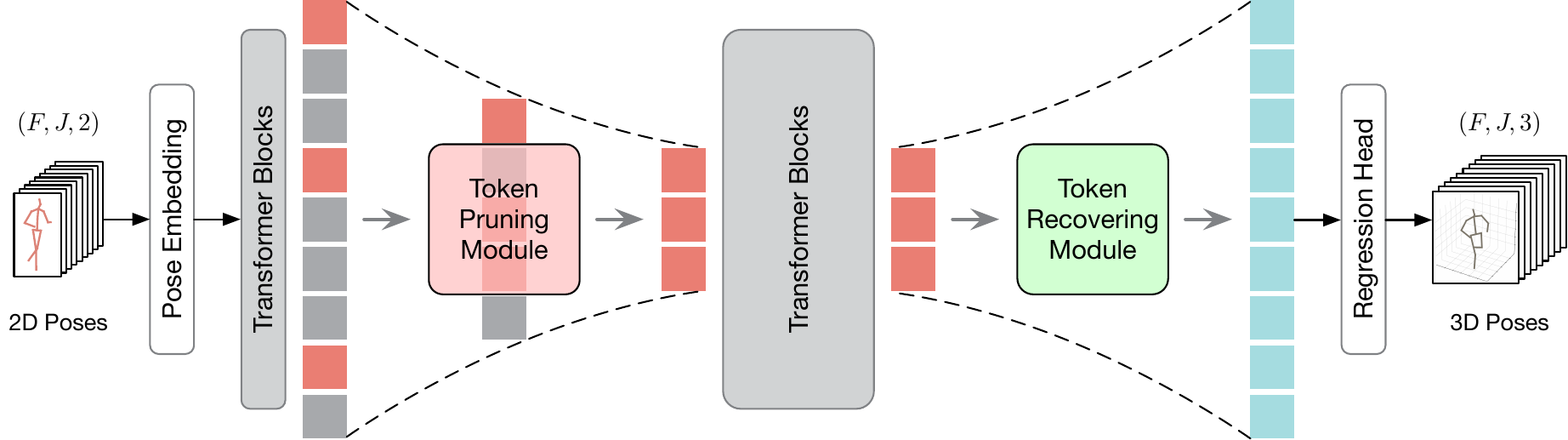}
\caption
{
Overview of the proposed Hierarchical Hourglass Tokenizer (H\textsubscript{2}OT).
H\textsubscript{2}OT mainly consists of a Token Pruning Module (TPM) and a Token Recovering Module (TRM). 
TPM selects the pose tokens of representative frames after the first few transformer blocks and TRM recovers the full-length tokens after the last transformer block. 
Note that TRM can be added either before or after the regression head and we add it before the regression head here. 
} 
\label{fig:overview}
\end{figure*}

Our method can be easily integrated into existing VPTs \cite{mhformer,mixste,motionbert,motionagformer} with minimal modifications (see Fig.~\ref{fig:overview}). 
Specifically, the first few transformer blocks of VPTs remain unchanged to obtain pose tokens with comprehensive information from full video frames. 
These pose tokens are then progressively pruned by the proposed TPM with its hierarchical pruning design. 
The remaining tokens, serving as representative tokens, are further fed into the subsequent transformer blocks. 
Finally, TRM recovers the full-length tokens after the last transformer block, while the intermediate transformer blocks still use representative tokens. 

To validate the effectiveness and efficiency of our method, we deploy it on top of SOTA VPTs, including MHFormer \cite{mhformer}, MixSTE \cite{mixste}, MotionBERT \cite{motionbert}, and MotionAGFormer \cite{motionagformer}. 
Extensive experiments demonstrate that the proposed method can greatly reduce the computational costs without sacrificing the pose estimation performance. 

This work can be summarized as follows:
\begin{itemize}
\item We present H\textsubscript{2}OT, a hierarchical plug-and-play \textit{pruning-and-recovering} framework for efficient transformer-based 3D HPE from videos. 
Our H\textsubscript{2}OT reveals that maintaining the full-length pose sequence is redundant, and identifying and keeping a few representative pose tokens can achieve both high efficiency and performance. 
\item To accelerate VPTs, we propose a Token Pruning Module (TPM) to select a few representative tokens for video redundancy reduction and a Token Recovering Module (TRM) to restore the original temporal resolution for fast inference. 
\item Extensive experiments conducted on four recent VPTs show that H\textsubscript{2}OT achieves highly competitive or even superior results while significantly improving efficiency. 
\end{itemize}

Part of the material presented in the paper appeared in our CVPR 2024 conference paper \cite{hot}. 
This journal version extends \cite{hot} in several ways:
\begin{itemize}
\item Unlike \cite{hot} that lacks a hierarchical design, we propose a novel hierarchical pruning strategy that gradually reduces the number of tokens as the layer gets deeper. 
Such a hierarchical design introduces a pyramidal feature hierarchy to VPTs, enabling our method to reduce video redundancy more effectively and further improve the efficiency of VPTs. 
\item We observe that cluster pruning and attention recovering strategies in \cite{hot} require additional parameters and impose a significant burden on the inference time compared to the original VPT models. 
To address this issue, we introduce sampling pruning and interpolation recovering strategies that are parameter-free and fast which makes H\textsubscript{2}OT more robust and suitable for real-world applications. 
\item With the above new designs, H\textsubscript{2}OT enables more efficient VPTs as compared with \cite{hot}, as validated in extensive ablation studies. 
\item We extend the experiments by comparing our H\textsubscript{2}OT with very recent works. 
Extensive experiments show that the proposed framework achieves both high efficiency and estimation accuracy compared with SOTA VPTs. 
\end{itemize}

\section{Related Work} 
\noindent \textbf{Transformer-based 3D HPE.}
Transformers are firstly proposed in \cite{transformer} have been successfully applied to different computer vision tasks \cite{ViT,swin,cheng2022gsrformer,tu2023implicit} and video-based 3D HPE \cite{mhformer,mixste,motionbert,shuai2022adaptive,zhang2025svtformer}. 
These video pose transformers (VPTs) are often built to capture spatial and temporal information for 3D HPE using transformers. 
PoseFormer \cite{poseformer} designs a transformer-based architecture to capture joint correlations and temporal dependencies. 
MHFormer \cite{mhformer} learns spatio-temporal multi-hypothesis representations of 3D human poses via transformers. 
MixSTE \cite{mixste} proposes a mixed spatio-temporal transformer to capture the temporal motion of different body joints. 
MotionBERT \cite{motionbert} presents a dual-stream spatio-temporal transformer to model long-range spatio-temporal relationships among skeletal joints. 

However, the improved performance of these VPTs comes with a heavy computation burden, which limits their applications in practice. 
In this work, we improve the efficiency of existing VPTs by keeping a few representative tokens in the intermediate transformer blocks. 

\noindent \textbf{Efficient 3D HPE.}
Efficient 3D HPE is critical in computing resource-constrained environments. 
Existing explorations mainly focus on efficient architecture design \cite{simplebaseline,videopose,choi2021mobilehumanpose} and data redundancy reduction \cite{stride,pstmo,deciwatch,einfalt2023uplift}. 
VPose \cite{videopose} presents a fully convolutional architecture that processes multiple frames in parallel. 
Strided \cite{stride} designs a strided transformer encoder to aggregate redundant sequences. 
Recently, several studies \cite{pstmo,deciwatch,einfalt2023uplift} have attempted to improve model efficiency from the input sequence.
For example, P-STMO \cite{pstmo} proposes a temporal downsampling strategy on the input side to diminish data redundancy. 
DeciWatch \cite{deciwatch} proposes a flow that takes sparsely sampled frames as inputs. 
Einfalt \textit{et al.} \cite{einfalt2023uplift} designs a joint uplifting and upsampling transformer architecture to generate dense 3D poses from sparse sequences of 2D poses. 
Besides, many efficient methods \cite{stride,einfalt2023uplift,poseformerv2} are designed for a specific model and some methods \cite{deciwatch,sun2023mixsynthformer} are built on single-frame estimators \cite{simplebaseline,spin,eft,pare}. 

However, none of them unifies the efficient design for different VPTs. 
We are the first to propose a plug-and-play framework for efficient VPTs, which can be plugged into common VPT models. 

\noindent \textbf{Token Pruning for Transformers.}
The self-attention complexity in transformers grows quadratically with the number of tokens, making it infeasible for high spatial or temporal resolution inputs.  
Several studies \cite{dou2023tore,long2023beyond,kong2022spvit,chang2023making} attempt to alleviate this issue with token pruning, aiming to select significant tokens from different inputs. 
The studies show that discarding less informative tokens in the deep transformer blocks only leads to a slight performance drop. 
DynamicViT \cite{rao2021dynamicvit} proposes a learnable prediction module to estimate the scores of tokens and prune redundant tokens. 
PPT \cite{ma2022ppt} selects important tokens based on the attention score. 
TCFormer \cite{zeng2024tcformer} presents a token clustering transformer to cluster and merge tokens. 
VTC-LFC \cite{wang2022vtc} compresses ViTs from frequency domain and prunes both parameters and tokens. 
GTPT \cite{wang2024gtpt} groups human keypoint tokens and prunes visual tokens to reduce redundancy. 

In this work, we are the first to perform token pruning in VPTs for model acceleration. 
Unlike these studies that aim to reduce less related information (\textit{e.g.}, image background) from images in the spatial domain, we focus on reducing video redundancy by selecting a few pose tokens of representative frames in the temporal domain. 
Furthermore, we propose to restore the full-length temporal resolution to meet the domain-specific requirement of efficient video-based 3D HPE. 

\section{Method}
Fig.~\ref{fig:overview} illustrates the overview of our Hierarchical Hourglass Tokenizer (H\textsubscript{2}OT). 
We propose a Token Pruning Module (TPM) and a Token Recovering Module (TRM) for token pruning and recovering. 
Our H\textsubscript{2}OT is a general-purpose \textit{pruning-and-recovering} framework, where TPM and TRM can use different token pruning and token recovering strategies, and we insert them into SOTA VPTs \cite{mhformer,mixste,motionbert,motionagformer}. 
In the following, we give details about the proposed TPM and TRM and show how to apply them to existing VPTs. 

\subsection{Token Pruning Module}

We observe that the existing VPTs \cite{{mhformer,mixste,motionbert,motionagformer}} take long video sequences as input and maintain the full-length sequence across all blocks (Fig.~\ref{fig:pipeline} (a)), which is computationally expensive for high temporal resolution inputs. 
To tackle this issue, we propose a Token Pruning Module (TPM) that prunes the pose tokens of video frames to improve the efficiency of VPTs. 
In our earlier conference work \cite{hot}, we maintain the full-length sequence in the early stages and perform token pruning in large chunks which often leads to clear information loss. 

Inspired by the hierarchical design in CNNs \cite{he2016deep,simonyan2014very}, we design hierarchical pruning that gradually reduces the number of tokens across network layers, leading to a pyramidal feature hierarchy of VPTs which helps preserve more useful information. 
Algorithm \ref{alg:hierarchical} shows the overall algorithm of the proposed hierarchical token pruning. 
Compared to the one-time pruning design (as in our published conference paper \cite{hot}) that maintains the full-length sequence in the early stages, this hierarchical pruning design reduces the sequence at various stages, creating a pyramidal feature hierarchy for VPTs (see Figure \ref{fig:pipeline}). 
Let $r_m \in\left\{r_1, r_2, \ldots, r_M\right\}$ be the number of representative tokens, $b_m \in\left\{b_1, b_2, \ldots, b_M\right\}$ be the block index of representative tokens, and $L$ be the number of transformer blocks at the $m$-th stage.
TPM takes the pose tokens $x_{b_m} \in \mathbb{R}^{r_{m-1} \times J \times C}$ of $b_m$-th transformer block as inputs and outputs a few representative tokens $\tilde{x}_{b_m} \in \mathbb{R}^{r_m \times J \times C}$ ($r_m < r_{m-1}$). 
Here, $J$ is the number of body joints, $C$ is the feature dimension, and $r_0=F$ is the number of input frames.
Note that our method only prunes the tokens along the temporal dimension since the frame number $F$ is much larger than the joint number $J$ (\textit{e.g.}, $F = 243$ and $J = 17$), \textit{i.e.}, the expensive and redundant computational costs are dominated by the frame number in the temporal domain. 

Another challenging question is how to select a few pose tokens that maintain rich information for accurate 3D HPE. 
In the following, we explore four token pruning strategies to answer this question, including Token Pruning Cluster (TPC), Token Pruning Attention (TPA), Token Pruning Motion (TPMo), and Token Pruning Sampler (TPS). 

\begin{algorithm}[t]
    \caption{Hierarchical Pruning Strategy}\label{alg:hierarchical}
    \textbf{Input}: Pose tokens $x \in \mathbb{R}^{F \times J \times C}$ \\
    \textbf{Output}: Representative tokens $\tilde{x} \in \mathbb{R}^{r_M \times J \times C}$ 
    \begin{algorithmic}[1]
    \For{layer index $l=1$ in $L$}
    \For{stage index $m=1$ in $M$}
    \If{$l$ = $b_{m}$}
    \State Token pruning at $b_{m}$-th transformer block
    \EndIf
    \EndFor
    \State Feature extraction at $l$-th transformer block 
    \EndFor
    \State \textbf{return} $\tilde{x}$
    \end{algorithmic}
\end{algorithm}

\begin{figure}[tb]
\centering
\includegraphics[width=1.00\linewidth]{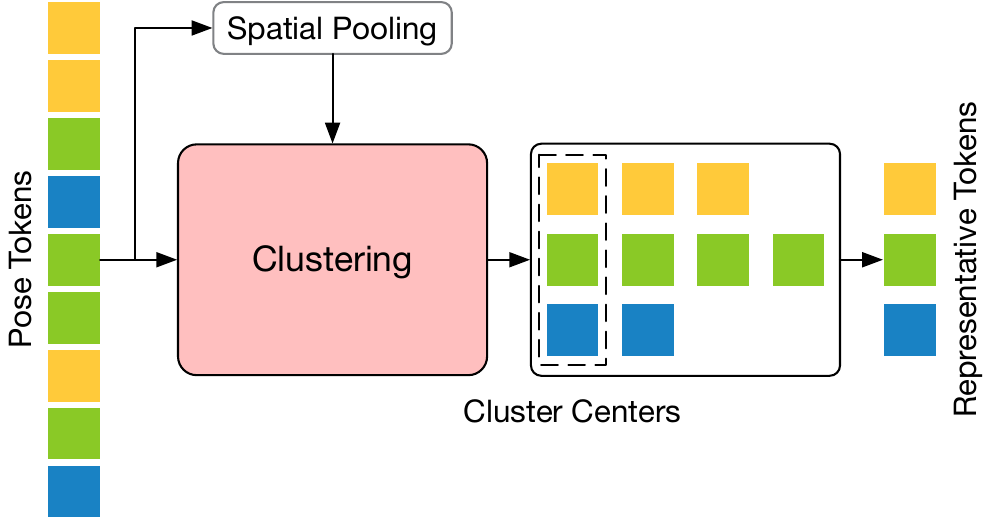}
\caption
{
Illustration of our Token Pruning Cluster (TPC) architecture. 
Given the input pose tokens, we pool them in the spatial dimension, cluster the input tokens into several groups
according to the feature similarity of the pooled tokens, and select the cluster centers as the representative tokens. 
}
\label{fig:tpc}
\end{figure}

\subsubsection{Token Pruning Cluster}

We propose a simple, effective, and parameter-free Token Pruning Cluster (TPC) that dynamically selects a few pose tokens of representative frames to eliminate video redundancy. 
The architecture of TPC is illustrated in Figure~\ref{fig:tpc}. 
Given the input pose tokens of the $b_m$-th transformer block $x_{b_m} \in \mathbb{R}^{r_{m-1} \times J \times C}$, an average spatial pooling is used along the spatial dimension to remove spatial redundancy, resulting in pooled tokens $\overline{x}_{b_m} \in \mathbb{R}^{r_{m-1} \times C}$. 
Then, we apply an efficient density peaks clustering based on $k$-nearest neighbors (DPC-$k$NN) algorithm \cite{du2016study}. 
This algorithm clusters the input pose tokens into several groups according to the feature similarity of the pooled tokens without requiring an iterative process. 

The cluster centers of tokens are characterized by a higher density compared to their neighbors, as well as a relatively large distance from other tokens with higher densities. 
For a token $x^{i} \in \overline{x}_{b_m}$, the local density of tokens $\rho$ is calculated by:
\begin{equation}
    \rho_{i}=\exp (-\frac{1}{k} \sum_{x^{j} \in k\mathrm{NN}({x}^{i})} 
    \left\|x^i-x^j\right\|_{2}^{2}),
\end{equation}
where $k\mathrm{NN}\left({x}^{i}\right)$ are the $k$-nearest neighbors of a token $x^{i}$. 

We then define the $\delta_{i}$ that measures the minimal distance between the token $x^{i}$ and other tokens with higher density. 
The $\delta_{i}$ of the token with the highest density is set to the maximum distance between it and any other tokens. 
The $\delta_{i}$ of each token is calculated by:
\begin{equation}
    \delta_{i}=\left\{
    \begin{array}{l}
    \min _{j: \rho_j>\rho_i} \left\|x^i-x^j\right\|_{2}, \text { if } \exists \rho_{j}>\rho_{i} \\
    \max _{j} \left\|x^i-x^j\right\|_{2}, \text { otherwise }
    \end{array}.\right.
\end{equation}

The clustering center score of a token $x^{i}$ is computed by multiplying the local density $\rho_i$ and the minimal distance $\delta_i$ (\textit{i.e.}, as $\rho_i \times \delta_i$). 
A higher score indicates that the token has a large density and distance, and so a higher potential to be the cluster center. 
The top-$r_m$-scored input pose tokens are selected as cluster centers, and the remaining tokens are assigned to the nearest cluster center with higher density.

The cluster centers have high semantic diversity, containing more informative information than the other tokens. 
Therefore, the cluster centers serve as the representative tokens $\tilde{x}_{b_m} \in \mathbb{R}^{r_m \times J \times C}$ for efficient estimation, and the remaining tokens are discarded for reduction of video redundancy. 

\subsubsection{Token Pruning Attention}

VPTs \cite{poseformer,stride,mhformer,mixste,motionbert} are based on the transformer architecture \cite{transformer}, which is effective at capturing long-range spatio-temporal dependencies.
The core of the transformer is the self-attention mechanism, which matches a query with a set of key-value pairs to produce an output \cite{transformer}. 
Given the input $x$, three linear projections are used to transform $x$ into three matrices: queries $Q$, keys $K$, and values $V$. 
The self-attention operation can then be computed as follows:
\begin{equation}
    \operatorname{Attention}(Q, K, V)=\operatorname{Softmax}\left(\frac{Q K^T}{\sqrt{d}}\right) V=\alpha V,
\label{equ:attention}
\end{equation}
where $Q \in \mathbb{R}^{n_{q} \times d}$, $K \in \mathbb{R}^{n_{k} \times d}$, and $V \in \mathbb{R}^{n_{v} \times d}$. 
$d$ is the dimension and $n_{q} = n_{k} = n_{v} = n$ is the number of tokens. 
$\alpha \in \mathbb{R}^{n \times n}$ is the attention score that determines how much information of each pose token is fused into the output. 
Thus, the attention score $\alpha$ can be used to select informative tokens. 

To this end, we propose a Token Pruning Attention (TPA) module that uses the attention score to select a few pose tokens of representative frames. 
Given the input pose tokens of the $b_m$-th transformer block $x_{b_m} \in \mathbb{R}^{r_{m-1} \times J \times C}$ and the attention score $\alpha_{b_m} \in \mathbb{R}^{r_{m-1} \times r_{m-1}}$, we use $\overline{\alpha}_{b_m} \in \mathbb{R}^{r_{m-1}} =\sum_j {\alpha}^{i,j}_{b_m}$ as the criterion and select the top-$r_m$-scored pose tokens as representative tokens $\tilde{x}_{b_m} \in \mathbb{R}^{r_m \times J \times C}$. 
This attention score based pruning strategy is learnable and can adaptively select the most informative tokens for efficient estimation. 

\subsubsection{Token Pruning Motion}

Human motion can reveal variations in body movement. 
Therefore, it is reasonable to assume that by using human motion to identify keyframes in a motion sequence, we can retain frames with significant changes while removing those with minimal changes, thereby reducing redundant information. 

With this assumption, we propose a Token Pruning Motion (TPMo) module that selects a few pose tokens of representative frames based on human motion. 
Specifically, given the input 2D pose sequence $p \in \mathbb{R}^{F \times J \times 2}$ detected by an off-the-shelf 2D HPE detector from a video, we first transform it into $s \in \mathbb{R}^{F \times (J \cdot 2)}$. 
For the $b_m$-th transformer block, the human motion can be formulated as:
\begin{equation}
    M_{b_m} =\left\{0, s_{b_m}^2-s_{b_m}^1, s_{b_m}^3-s_{b_m}^2, \cdots, s_{b_m}^{t+1}-s_{b_m}^{t}\right\},
\end{equation}
where $s_{b_m}^t$ is the $t$-th frame of the sequence $s$ in the $b_m$-th transformer block, and $M_{b_m} \in \mathbb{R}^{r_{m-1} \times (J \cdot 2)}$ is the motion matrix. 
Then, we use $\overline{M}_{b_m} \in \mathbb{R}^{r_{m-1}} =\sum_j M_{b_m}^{i,j}$ as the criterion and select the top-$r_m$-scored pose tokens as representative tokens $\tilde{x}_{b_m} \in \mathbb{R}^{r_m \times J \times C}$. 
This human motion based pruning strategy can consider redundancy and context simultaneously to select frames with a large variation in motion. 

\begin{figure}[tb]
\centering
\includegraphics[width=1.00\linewidth]{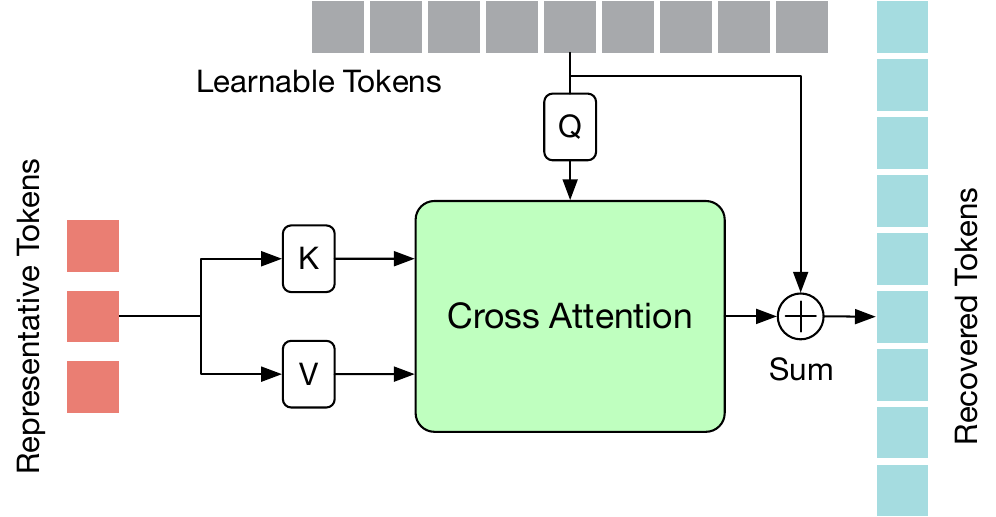}
\caption
{
Illustration of our Token Recovering Attention (TRA). 
TRA takes the representative tokens of the last transformer block, along with learnable tokens that are initialized to zero, as input to recover the full-length tokens. 
}
\label{fig:tra}
\end{figure}

\subsubsection{Token Pruning Sampler}

Our conference version \cite{hot} chooses the TPC module for token pruning, which is simple, effective, and parameter-free. 
However, we observe some disadvantages in TPC used in \cite{hot}:
\textbf{(i)} TPC utilizes a cluster-based algorithm for token pruning, but it consumes a big burden on the inference time compared to the original VPT models. 
\textbf{(ii)} Employing clustering necessitates token selection in an unordered manner which clearly degrades the interpolation efficiency. 
This issue also exists in the dynamic token pruning strategies like TPA and TPMo. 

To address these issues, we propose a Token Pruning Sampler (TPS) module that uses a linear sampling strategy to select a few pose tokens. 
Given the input pose tokens of $b_m$-th transformer block $x_{b_m} \in \mathbb{R}^{r_{m-1} \times J \times C}$, TPS uniformly samples pose tokens along the temporal dimension and output a few representative tokens $\tilde{x}_{b_m} \in \mathbb{R}^{r_m \times J \times C}$.
Such a simple sampling strategy is reasonable due to data redundancy (i.e., nearby poses are similar) on Human3.6M dataset \cite{ionescu2013human3} (captured by 50 Hz cameras). 
TPS is parameter-free, efficient, and requires no additional computational costs and inference time. 
Besides, it enables the token selection in an ordered manner, which is more suitable for efficient interpolation. 

\subsection{Token Recovering Module}

A large number of pose tokens have been pruned by TPM, which significantly reduces the computational costs. 
However, for fast inference, a real-world 3D HPE system should be capable of estimating the consecutive 3D poses of all frames in a given video (this is called \textit{seq2seq} pipeline in \cite{mixste}). 
Therefore, different from some token pruning methods in vision transformers that can use a few selected tokens to directly perform classification \cite{yin2022vit,marin2023token,liang2022not,chen2023diffrate}, we need to recover the full-length tokens to keep the same number of tokens as the input video frames (in existing VPTs, each token corresponds to a frame). 
Meanwhile, for efficiency purposes, the recovering module should be lightweight. 

\subsubsection{Token Recovering Attention}

We develop a lightweight Token Recovering Attention (TRA) module to restore the spatio-temporal information from the selected pose tokens, as shown in Fig.~\ref{fig:tra}. 
It only contains one multi-head cross-attention (MCA) layer without any additional networks. 
The dot-product attention \cite{transformer} in the MCA is defined in Eq.~\ref{equ:attention}. 

Our MCA takes the learnable tokens $x^{\prime} \in \mathbb{R}^{F \times C}$ that are initialized to zero as queries and the $j$-th joint representative tokens of the last transformer block $x_{L}^{j} \in \mathbb{R}^{r_M \times C}$ as keys and values, followed by a residual connection:
\begin{equation}
    \hat{x}^{j} = 
    x^{\prime} + 
    \operatorname{MCA}(x^{\prime}, x_{L}^{j}, x_{L}^{j}),
\end{equation}
where $L$ is the number of transformer blocks. $\operatorname{MCA}(\cdot)$ is the function of MCA, and its inputs are queries, keys, and values. 
$\hat{x}^{j} \in \mathbb{R}^{F \times C}$ is the $j$-th joint recovered token, whose temporal dimension is the same as the queries (\textit{i.e.}, the designed learnable tokens). 

The TRA performs a reverse operation of selecting representative tokens, which recovers tokens of full-length temporal resolution from low ones using high-level spatio-temporal semantic information. 

\subsubsection{Token Recovering Interpolation}

Although TRA is an effective design choice for token recovering, it still requires additional parameters and computational costs. 
To address this issue in our conference paper \cite{hot}, we design a Token Recovering Interpolation (TRI) module that uses a simple interpolation operation to recover the full-length 3D human poses. 
The tokens of the last transformer block $x_{L} \in \mathbb{R}^{r_M \times J \times C}$ are fed into a regression head to estimate the 3D pose sequence $\hat{q} \in \mathbb{R}^{r_M \times J \times 3}$. 
Then, a linear interpolation operation is used to recover the 3D poses of all frames $q \in \mathbb{R}^{F \times J \times 3}$. 
Note that efficient interpolation with parallel computing requires that the interpolated sequence should be ordered. Therefore, in TPM, it is necessary to use TPS in conjunction with TRI. 

\subsection{Applying to Video Pose Transformers}
\label{sec:VPTs}

\subsubsection{Video Pose Transformers}

\begin{figure}[t]
\centering
\includegraphics[width=1.00\linewidth]{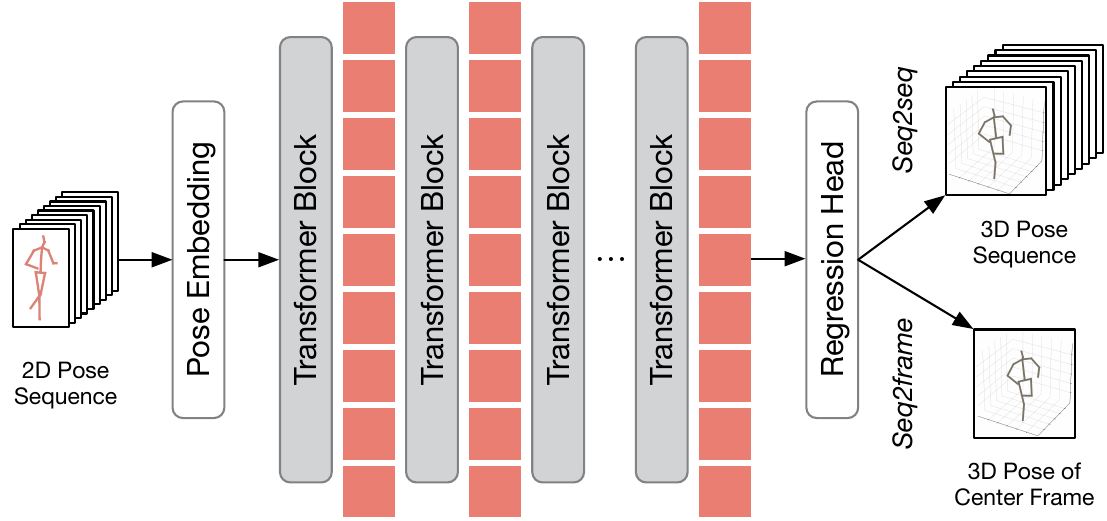}
\caption
{
Summary of existing VPT architectures. 
Existing VPTs typically contain a pose embedding module, a stack of transformer blocks, and a regression head module. 
The outputs of the regression head can be either the 3D poses of all frames for the \textit{seq2seq} pipeline or the 3D pose of the center frame for the \textit{seq2frame} pipeline.
}
\label{fig:vpt_supp}
\end{figure}

Recent studies of video pose transformers (VPTs) \cite{poseformer,stride,li2023multi,mixste,pstmo,motionbert} are mainly designed to estimate 3D poses from 2D pose sequences. 
These VPTs share a similar architecture, which includes a pose embedding module (often containing only a linear layer) to embed spatial and temporal information of pose sequences, a stack of transformer blocks to learn global spatio-temporal correlations, and a regression module to predict 3D human poses. 
We summarize the architecture in Fig.~\ref{fig:vpt_supp}. 

Each transformer block consists of a multi-head self-attention (MSA) layer and a feed-forward network (FFN) layer. 
Let $N$ be the number of tokens, $D$ be the dimension, and $2D$ be the expanding dimension in the FFN (the expanding ratio in VPTs is typically 2). 
The calculational costs of MSA and FFN are $\mathcal{O}\left(4 N D^2+2 N^2 D\right)$, and $\mathcal{O}\left(4 N D^2\right)$, respectively. 
Thus, the total computational complexity is $\mathcal{O}\left(8 N D^2+2 N^2 D\right)$, which makes VPTs computationally expensive. 
Since the dimension $D$ is important to determine the modeling ability and most recent VPTs employ a $D$ of 512 or 256, we follow their hyperparameter settings and propose to prune pose tokens of video frames (\textit{i.e.}, reducing $N$) to reduce the computational cost of VPTs. 

There are two types of pipelines based on VPT's inference outputs: \textit{seq2frame} \cite{poseformer,mhformer,stride,pstmo} and \textit{seq2seq} \cite{mixste,motionbert,motionagformer} pipelines. 
The \textit{seq2frame} pipeline outputs the 3D pose of the center frame and requires repeated inputs of 2D pose sequences with significant overlap to predict the 3D poses of all frames, which makes it less efficient but can achieve better performance by considering both past and future information. 
In contrast, the \textit{seq2seq} pipeline outputs 3D poses of all frames from the input 2D pose sequence at once, making it more efficient but less accurate. 
As a result, these two pipelines have their unique strengths, and we need to develop two strategies to better accommodate their different inference manners. 

\subsubsection{Applying to \textit{Seq2seq} Pipeline}

For the \textit{seq2seq} pipeline, the outputs are all frames of the input video, and hence we need to restore the original temporal resolution. 
TPM and TRM are inserted into VPTs, where TPM prunes the tokens after a few transformer blocks and TRM recovers the full-length tokens after the last transformer block, as shown in Fig.~\ref{fig:overview}. 
Given the input 2D pose sequence $p \in \mathbb{R}^{F \times J \times 2}$ detected by an off-the-shelf 2D HPE detector from a video, we first feed $p$ into a pose embedding module to embed spatial and temporal information, resulting in tokens $x \in \mathbb{R}^{F \times J \times C}$. 
The embedded tokens are then fed into first few transformer blocks. 
Next, TPM selects representative tokens $\tilde{x} \in \mathbb{R}^{r_M \times J \times C}$ after several hierarchical pruning stages, and use them as the inputs of subsequent transformer blocks. 
TRM has two different design choices: TRA and TRI. 
\textbf{(i)} TRA recovers the full-length tokens from the tokens of the last transformer block $x_{L} \in \mathbb{R}^{r_M \times J \times C}$, resulting in recovered tokens $\hat{x} \in \mathbb{R}^{F \times J \times C}$. 
It finally includes a regression head to estimate the 3D poses of all frames $q \in \mathbb{R}^{F \times J \times 3}$. 
\textbf{(ii)} For TRI, the tokens of the last transformer block $x_{L} \in \mathbb{R}^{r_M \times J \times C}$ are fed into the regression head to estimate the 3D pose sequence $\hat{q} \in \mathbb{R}^{r_M \times J \times 3}$. 
Then TRI utilizes the interpolation operation to recover the 3D poses of all frames $q \in \mathbb{R}^{F \times J \times 3}$. 

\begin{figure}[t]
\centering
\includegraphics[width=1.00\linewidth]{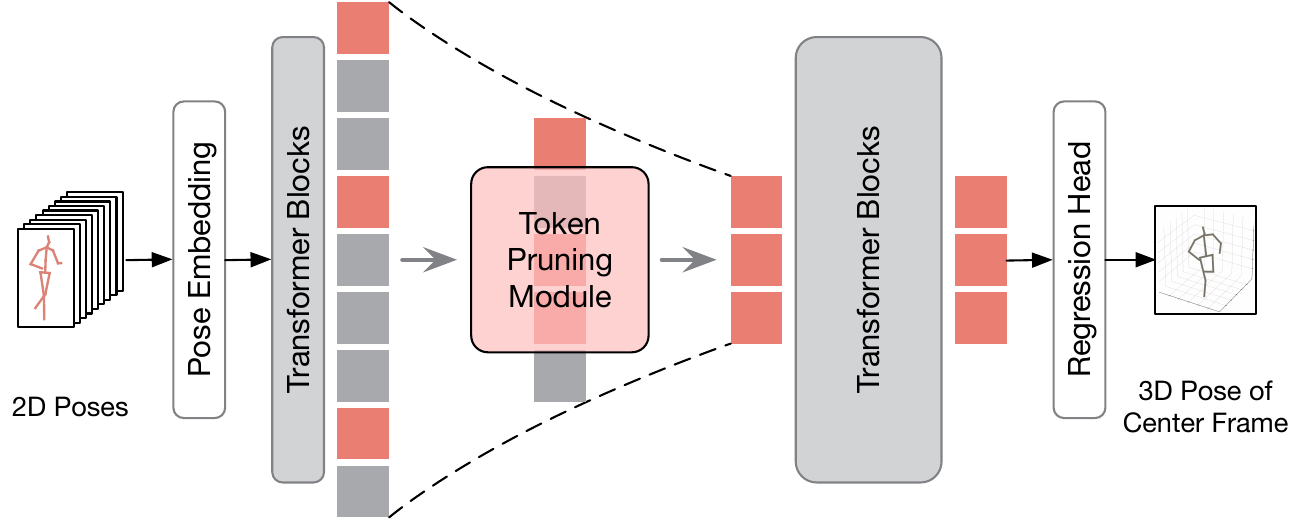}
\caption
{
Illustration of our framework on \textit{seq2frame} pipeline. 
The pose tokens are fed into TPM to select representative tokens. 
After the regression head, the 3D pose of the center frame is selected as the output for evaluation. 
}
\label{fig:seq2frame}
\end{figure}

\subsubsection{Applying to \textit{Seq2frame} Pipeline}

For the \textit{seq2frame} pipeline, the output is the 3D pose of the center frame. 
Therefore, TRM is unnecessary and we only insert TPM into VPTs. 
Since the token of the center frame directly corresponds to the output and can provide crucial information to the final estimation, we concatenate it with the selected tokens to make this pipeline work better. 
As shown in Fig.~\ref{fig:seq2frame}, the early stages of both pipelines share the same workflow. 
After the last transformer block, the tokens are directly sent to the regression head to perform regression and the 3D pose of center frame $q_{center} \in \mathbb{R}^{1 \times J \times 3}$ is selected as the final prediction. 

\section{Experiments}
\subsection{Datasets and Evaluation Metrics}
\noindent \textbf{Datasets.}
We evaluate our method on two 3D HPE benchmark datasets: 
Human3.6M \cite{ionescu2013human3} and MPI-INF-3DHP \cite{mehta2017monocular}. 
Human3.6M is the most widely used dataset for 3D HPE. 
It consists of 3.6 million video frames recorded by four RGB cameras at 50 Hz in an indoor environment. 
This dataset includes 11 actors performing 15 daily actions. 
Following \cite{li2025graphmlp,li2023multi,ci2020locally,xu2021monocular}, subjects S1, S5, S6, S7, S8 are used for training and subjects S9, S11 are used for testing. 
MPI-INF-3DHP is another popular 3D HPE dataset. 
This dataset contains 1.3 million frames collected in indoor and outdoor scenes. 
It is smaller than Human3.6M but more challenging due to its diverse scenes, viewpoints, and motions. 

\noindent \textbf{Evaluation Metrics.}
For Human3.6M, we use the most commonly used mean per joint position error (MPJPE) as the evaluation metric, which measures the average Euclidean distance between estimated and ground truth 3D joint coordinates in millimeters. 
For MPI-INF-3DHP, we follow previous works \cite{poseformer,mhformer,mixste} to report metrics of MPJPE, percentage of correct keypoint (PCK) with the threshold of 150mm, and area under curve (AUC). 

\subsection{Implementation Details}
\label{sec:Implementation Details}

The network is implemented using the PyTorch framework on one consumer-level NVIDIA RTX 3090 GPU with 24G memory. 
Our method builds upon MHFormer \cite{mhformer}, MixSTE \cite{mixste}, MotionBERT \cite{motionbert}, and MotionAGFormer \cite{motionagformer} for their largest frame number (\textit{i.e.}, $F = 351,243,243,243$) models. 
We adopt the same optimal hyperparameters and training strategies used in \cite{mhformer,mixste,motionbert,motionagformer}, as detailed in Table \ref{table:details}. 
We also use the same loss functions for training, such as MPJPE loss for MHFormer, and weighted MPJPE loss, temporal consistency loss (TCLoss), and mean per-joint velocity error (MPJVE) for MixSTE. 
By default, we set \{$F = 351$, $r = [175, 117]$, $b = [0, 1]$\} for MHFormer, \{$F = 243$, $r = [121, 81]$, $b = [0, 3]$\} for MixSTE, \{$F = 243$, $r = [121, 81]$, $b = [0, 1]$\} for MotionBERT, and \{$F = 243$, $r = [121, 81]$, $b = [0, 7]$\} for MotionAGFormer. 
Note that MHFormer is designed for \textit{seq2frame} pipeline, so we only implement the TPM on it. 
MixSTE, MotionBERT, and MotionAGFormer are designed for \textit{seq2seq} pipeline and can be implemented on both \textit{seq2frame} (with TPM) and \textit{seq2seq} (with H\textsubscript{2}OT) pipelines.

\begin{table}[t]
\footnotesize
\centering
\caption
{ 
    Implementation details of our method on MHFormer \cite{mhformer}, MixSTE \cite{mixste}, MotionBERT \cite{motionbert}, and MotionAGFormer \cite{motionagformer}. 
    ($L$) - number of transformer blocks, 
    ($C$) - dimension, 
    (LR) - initial learning rate,
    (Flip) - horizontal flip augmentation,
    (CPN) - Cascaded Pyramid Network \cite{li2020cascaded},
    (SH) - Stack Hourglass \cite{newell2016stacked}. 
}
\setlength{\tabcolsep}{0.0mm}
\resizebox{\columnwidth}{!}{
\begin{tabular}{l|ccccc}
\toprule [1pt]
Config &MHFormer \cite{mhformer} &MixSTE \cite{mixste} &MotionBERT \cite{motionbert} & MotinAGFormer \cite{motionagformer} \\
\midrule [0.5pt]

$L$ &3 &8 &5 &16 \\

$C$ &512 &512 &256 &128 \\

Training Epoch &20 &160 &120 &120 \\

Batch Size &210 &4 &4 &4 \\

LR &$1 {\times} 10^{-3}$ &$4 {\times} 10^{-5}$ &$5 {\times} 10^{-4}$ &$5 {\times} 10^{-4}$ \\

Optimizer &Amsgrad &Adam &Adam &Adam \\

Augmentation &Flip &Flip &Flip &Flip  \\

2D Detector &CPN &CPN &SH &SH \\

\bottomrule [1pt]
\end{tabular}
}
\label{table:details}
\end{table}

\begin{table*}[t]
\footnotesize
\centering
\caption
{ 
    Ablation study on the different combinations of token pruning and recovering strategies under \textit{seq2seq} pipeline. 
``FPS/Frame'' denotes frame per second per each output frame. 
``FN'' denotes the frame noise that calculates the MPJPE of selected frames. 
``Ratio'' denotes the ratio of time spent on token pruning and recovering with the overall inference time. 
}
\setlength{\tabcolsep}{2.35mm}
\begin{tabular}{lccccccc|cccc|c}
\toprule [1pt]
& & \multicolumn{4}{c}{Token Pruning} & \multicolumn{2}{l}{Token Recovering} \\
\cmidrule(lr){3-6}\cmidrule(lr){7-8}
Method &FN &TPC &TPA &TPMo &TPS &TRA &TRI &Param (M) &FLOPs (G) &FPS/Frame &Ratio &MPJPE $\downarrow$ \\
\midrule [0.5pt]

MixSTE \cite{mixste} &6.61 & & & & & & &33.78 &277.25 &41 &- &40.9 \\ 
\midrule [0.5pt]

Ours &6.63 &\cmark & & & &\cmark &  &35.00 &167.52 &56 &0.33 &41.0 \\ 

Ours &6.56 & &\cmark & & &\cmark &  &35.00 &167.52 &63 &0.04 &42.1 \\ 

Ours &7.00 & & &\cmark & &\cmark & &35.00 &167.52 &62 &0.03 &42.8 \\

Ours &6.61 & & & &\cmark &\cmark &  &35.00 &167.52 &64 &0.03 &41.4 \\ 

Ours &6.61 & & & &\cmark & &\cmark &\textbf{\red{33.78}} &\textbf{\red{161.73}} &\textbf{\red{66}} &\textbf{\red{0.00}} &41.1 \\ 

\bottomrule [1pt]
\end{tabular}
\label{table:pruning and Recovering}
\end{table*}
    
\begin{figure*}[tb]
\centering
\includegraphics[width=1.00\linewidth]{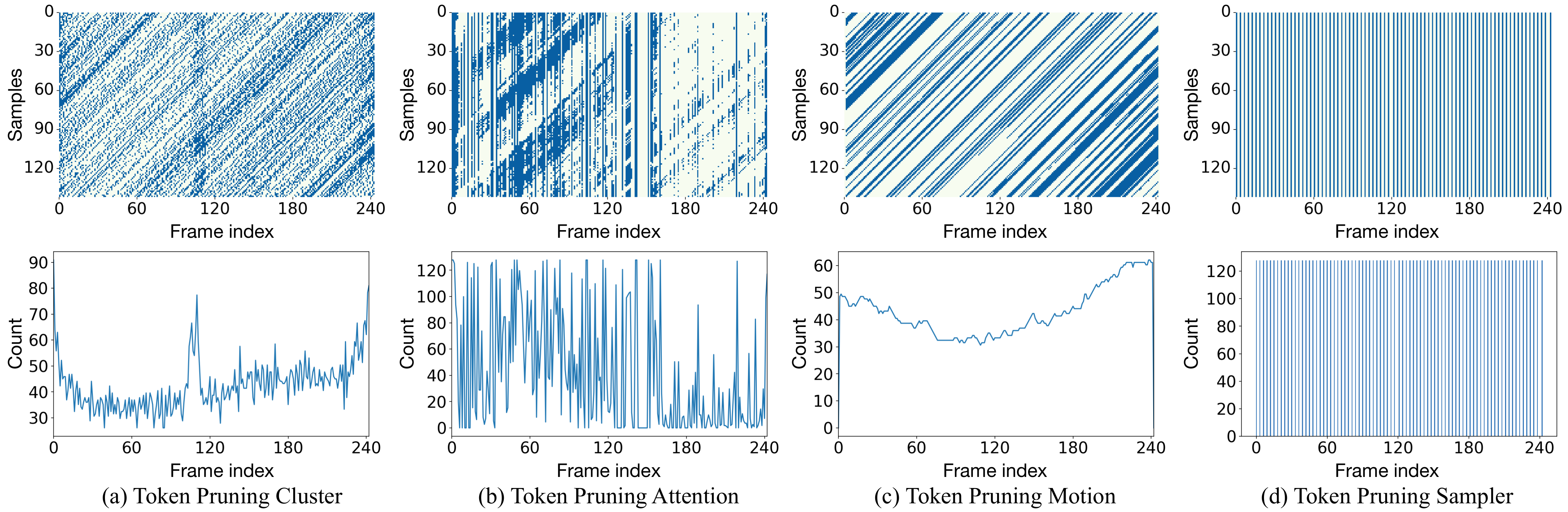}
\caption
{
Statistics visualization of selected tokens for different token pruning strategies. 
\textbf{Top}: Frame indexes of selected tokens for some samples (140 samples) of consecutive video sequences (243 frames). 
Blue points are selected tokens and white points are pruned tokens. 
\textbf{Bottom}: Frequency count of frame indexes of selected tokens for these samples. 
}
\label{fig:index}
\end{figure*}

\subsection{Ablation Study}
\label{sec:ablation}
To validate the effectiveness of our method, we conduct extensive ablation studies on Human3.6M dataset. 

\noindent \textbf{Token Pruning and Recovering Designs.}
Our method is a general-purpose \textit{pruning-and-recovering} framework that can be directly inserted into various token pruning and recovering techniques. 
In Table~\ref{table:pruning and Recovering}, we compare different combinations of TPC, TPA, TPMo, and TPS for token pruning, and TRA and TRI for token recovering. 
The hyper-parameters are all set by $F = 243$, $r = 81$, and $b = 3$ for a fair comparison. 
Since efficient interpolation requires the sequence to be ordered, TRI can only be combined with TPS. 
To measure the quality of selected tokens, we define a frame noise metric that calculates the MPJPE of the 2D poses of input frames corresponding to the selected indexes. 
As the table shows, the frame noises across these methods are similar (around 6.6mm) except for the motion pruning (7.0mm). 
This is because selecting tokens with top-$k$ large motion introduces noisy frames that differ significantly from clean frames, which can adversely affect performance. 
Moreover, the studies show that the combination of TPS and TRI requires the fewest parameters (33.78 M) and FLOPs (161.73 G) but achieves the highest FPS per frame (66). 
It also has the lowest relative ratio of time spent on token pruning and recovery (0.00) and achieves competitive performance (41.1 mm). 
As a comparison, TPC consumes more inference time (56 FPS/Frame and 0.33 Ratio) and TRA requires additional parameters and computational costs (35.00M Param and 167.52G FLOPs), while TPS and TRI are clearly more efficient. 
We therefore adopt TPS and TRI to create more powerful and faster VPT models. 

Furthermore, we statistically visualize selected tokens of four token pruning strategies: TPC, TPA, TPMo, and TPS. 
For better observation, we take samples of consecutive video sequences as input with a temporal interval of 1 between neighboring samples. 
The frame indexes and the frequency count of frame indexes of the selected tokens are shown in Figure~\ref{fig:index} (top) and Figure~\ref{fig:index} (bottom). 
TPS and TPMo are static pruning methods because the former selects tokens at a fixed frame interval (equidistance in the top of Figure~\ref{fig:index} (d)), while the latter selects tokens with top-$k$ large motions that move with the input sequence (oblique triangle in the top of Figure~\ref{fig:index} (c)). 
Instead, TPA and TPC are dynamic methods that consider the significance of input tokens. 
The bottom of Figure~\ref{fig:index}~(b) shows that TPA tends to select tokens in the left half of a sequence, indicating that the selected tokens tend to be similar to each other \cite{wang2022vtc}. 
TPC primarily selects tokens at the beginning, center, and end of a sequence (the bottom of Figure~\ref{fig:index}~(a)). 
This is reasonable since these three parts can represent the rough motion of an entire sequence, which contributes a lot to accurate estimation. 
These findings highlight that these different token pruning strategies have different characteristics, and all of them can select representative pose tokens to eliminate the redundancy of video frames. 

\begin{table*}[t]
\scriptsize
\centering
\caption
{ 
Comparison of efficiency and accuracy between \textit{seq2seq} ($*$) and \textit{seq2frame} ($\dagger$) inference pipelines. 
FPS, GPU memory cost (G), and training time (min/epoch) are computed on a single GeForce RTX 3090 GPU. 
}
\setlength{\tabcolsep}{4.05mm}
\begin{tabular}{l|cllll|c}
\toprule [1pt]
Method &Param (M) &FLOPs (G) &FPS &GPU Memory &Training Time &MPJPE $\downarrow$  \\

\midrule [0.5pt]

MixSTE \cite{mixste} ($*$) &33.78 &277.25 &9982 &11.4 &17.9 &40.9 \\

HoT w. MixSTE \cite{hot} ($*$) &35.00 &167.52 (\textbf{\blue{$\downarrow$ 39.6\%}}) & 14297 (\textbf{\blue{$\uparrow$ 43.2\%}}) &\ \ 7.6 (\textbf{\blue{$\downarrow$ 33.3\%}}) &11.2 (\textbf{\blue{$\downarrow$ 37.4\%}}) &{41.0} \\

\blue{H\textsubscript{2}OT} w. MixSTE (\textbf{Ours}) ($*$) &33.78 &118.23 (\textbf{\blue{$\downarrow$ 57.4\%}}) &18745 (\textbf{\blue{$\uparrow$ 87.8\%}}) &\ \ 5.6 (\textbf{\blue{$\downarrow$ 52.3\%}}) &\ \ 8.4 (\textbf{\blue{$\downarrow$ 53.1\%}}) &{40.5} \\

\midrule [0.5pt]

MixSTE \cite{mixste} ($\dagger$)  &33.78 &277.25 &41 &11.4 &17.9 &40.7 \\

TPC w. MixSTE \cite{hot} ($\dagger$) &33.78 &161.73 (\textbf{\blue{$\downarrow$ 41.7\%}}) &61 (\textbf{\blue{$\uparrow$ 48.8\%}}) &\ \ 7.3 (\textbf{\blue{$\downarrow$ 36.9\%}}) &10.7 (\textbf{\blue{$\downarrow$ 40.2\%}}) &{40.4} \\

\blue{TPM} w. MixSTE (\textbf{Ours}) ($\dagger$) &33.78 & 118.23 (\textbf{\blue{$\downarrow$ 57.4\%}}) &69 (\textbf{\blue{$\uparrow$ 68.3\%}}) &\ \ 5.6 (\textbf{\blue{$\downarrow$ 50.9\%}}) &\ \ 8.7 (\textbf{\blue{$\downarrow$ 51.4\%}})  &40.4 \\

\midrule [0.5pt]
\midrule [0.5pt]
MotionBERT \cite{motionbert} ($*$) &16.00 &131.09 &12719 &10.7 &18.5 &39.8 \\

HoT w. MotionBERT \cite{hot} ($*$) &16.35 &\ \ 63.21 (\textbf{\blue{$\downarrow$ 51.8\%}}) &15289 (\textbf{\blue{$\uparrow$ 20.2\%}}) &\ \ 6.1 (\textbf{\blue{$\downarrow$ 43.0\%}}) &11.5 (\textbf{\blue{$\downarrow$ 37.8\%}}) &{39.8} \\

\blue{H\textsubscript{2}OT} w. MotionBERT (\textbf{Ours}) ($*$) &16.00 &\ \ 47.98 (\textbf{\blue{$\downarrow$ 63.4\%}}) & 18779 (\textbf{\blue{$\uparrow$ 47.6\%}}) &\ \ 4.4 (\textbf{\blue{$\downarrow$ 58.9\%}}) &\ \ 9.0 (\textbf{\blue{$\downarrow$ 51.4\%}})  &39.9 \\

\midrule [0.5pt]

MotionBERT \cite{motionbert} ($\dagger$) &16.00 &131.09 &52 &10.7 &18.5 &39.5 \\

TPC w. MotionBERT \cite{hot}
($\dagger$) &16.00 &\ \ 61.04 (\textbf{\blue{$\downarrow$ 53.4\%}}) &68 (\textbf{\blue{$\uparrow$ 30.8\%}}) &\ \ 5.7 (\textbf{\blue{$\downarrow$ 46.7\%}}) & 10.9 (\textbf{\blue{$\downarrow$ 41.1\%}})  &{39.2} \\

\blue{TPM} w. MotionBERT (\textbf{Ours}) ($\dagger$) &16.00 &\ \ 47.98 (\textbf{\blue{$\downarrow$ 63.4\%}}) & 76 (\textbf{\blue{$\uparrow$ 46.2\%}}) &\ \ 4.4 (\textbf{\blue{$\downarrow$ 58.9\%}}) &\ \ 8.9 (\textbf{\blue{$\downarrow$ 51.9\%}})  &39.3 \\

\bottomrule [1pt]

\end{tabular}
\label{table:pipeline}
\end{table*}

\begin{table*}[t]
\scriptsize
\centering
\caption
{ 
Ablation studies on the number of representative tokens ($r$) and the block index of representative tokens ($b$) under \textit{seq2seq} pipeline. 
Frame per second
(FPS) was computed on a single GeForce RTX 3090 GPU. 
}
\setlength{\tabcolsep}{2.50mm}
\begin{tabular}{l|llcllll|cc}
\toprule [1pt]
Method &$r$ &$b$  &Param (M) &FLOPs (G) &FPS/Frame &GPU Memory &Training Time &MPJPE $\downarrow$  \\

\midrule [0.5pt]
\midrule [0.5pt]

MixSTE \cite{mixste} &243 &- &33.78 &277.25 &41  &11.4 &17.9 &40.9 \\

HoT w. MixSTE \cite{hot} &81 &3  &35.00 &167.52 (\textbf{\blue{$\downarrow$ 39.6\%}}) &56 (\textbf{\blue{$\uparrow$ 36.6\%}})  &7.6 (\textbf{\blue{$\downarrow$ 33.3\%}}) &11.2 (\textbf{\blue{$\downarrow$ 37.4\%}}) &{41.0} \\

\midrule [0.5pt]

\blue{H\textsubscript{2}OT} w. MixSTE &[81] &[3] &33.78 &161.73 (\textbf{\blue{$\downarrow$ 41.7\%}}) &66 (\textbf{\blue{$\uparrow$ 61.0\%}}) &7.3 (\textbf{\blue{$\downarrow$ 36.0\%}}) &10.7 (\textbf{\blue{$\downarrow$ 40.2\%}}) &41.0 \\

\blue{H\textsubscript{2}OT} w. MixSTE &[121, 81] &[0, 3] &33.78 &118.23 (\textbf{\blue{$\downarrow$ 57.4\%}}) &82 (\textbf{\blue{$\uparrow$ 100.0\%}}) &5.6 (\textbf{\blue{$\downarrow$ 50.9\%}}) & \ \ 8.0 (\textbf{\blue{$\downarrow$ 55.3\%}}) &\textbf{\red{40.5}} \\

\blue{H\textsubscript{2}OT} w. MixSTE &[121, 81, 49] &[0, 3, 5] &33.78 &104.54 (\textbf{\blue{$\downarrow$ 62.3\%}}) &85 (\textbf{\blue{$\uparrow$ 107.3\%}}) &5.1 (\textbf{\blue{$\downarrow$ 55.3\%}}) & \ \ 7.9 (\textbf{\blue{$\downarrow$ 55.9\%}}) &40.7 \\

\blue{H\textsubscript{2}OT} w. MixSTE &[189, 121, 81, 49] &[0, 3, 5, 7] &33.78 &149.32 (\textbf{\blue{$\downarrow$ 46.1\%}}) &74 (\textbf{\blue{$\uparrow$ 80.5\%}}) &6.7 (\textbf{\blue{$\downarrow$ 41.2\%}}) &10.1 (\textbf{\blue{$\downarrow$ 43.6\%}}) &40.8 \\

\midrule [0.5pt]
\midrule [0.5pt]    

\blue{H\textsubscript{2}OT} w. MixSTE &[189, 81] &[0, 3] &33.78 & 142.48 (\textbf{\blue{$\downarrow$ 48.6\%}}) &75 (\textbf{\blue{$\uparrow$ 82.9\%}}) &6.4 (\textbf{\blue{$\downarrow$ 43.9\%}}) &10.2 (\textbf{\blue{$\downarrow$ 43.0\%}}) &40.7 \\

\blue{H\textsubscript{2}OT} w. MixSTE &[121, 81] &[0, 3] &33.78 & 118.23 (\textbf{\blue{$\downarrow$ 57.4\%}}) &82 (\textbf{\blue{$\uparrow$ 100.0\%}}) &5.6 (\textbf{\blue{$\downarrow$ 50.9\%}}) & \ \ 8.0 (\textbf{\blue{$\downarrow$ 55.3\%}}) &\textbf{\red{40.5}} \\

\blue{H\textsubscript{2}OT} w. MixSTE &[121, 49] &[0, 3] &33.78 & \ \ 95.41 (\textbf{\blue{$\downarrow$ 65.6\%}}) &82 (\textbf{\blue{$\uparrow$ 100.0\%}}) &4.8 (\textbf{\blue{$\downarrow$ 57.9\%}}) & \ \ 6.8 (\textbf{\blue{$\downarrow$ 62.0\%}}) &41.2 \\

\blue{H\textsubscript{2}OT} w. MixSTE &[121, 27] &[0, 3] &33.78 & \ \ 79.73 (\textbf{\blue{$\downarrow$ 71.2\%}}) &83 (\textbf{\blue{$\uparrow$ 102.4\%}}) &4.4 (\textbf{\blue{$\downarrow$ 61.4\%}}) & \ \ 6.5 (\textbf{\blue{$\downarrow$ 63.7\%}}) &41.3 \\

\midrule [0.5pt]

\blue{H\textsubscript{2}OT} w. MixSTE &[121, 81] &[0, 3] &33.78 & 118.23 (\textbf{\blue{$\downarrow$ 57.4\%}}) &82 (\textbf{\blue{$\uparrow$ 100.0\%}}) &5.6 (\textbf{\blue{$\downarrow$ 50.9\%}}) & \ \ 8.0 (\textbf{\blue{$\downarrow$ 55.3\%}}) &\textbf{\red{40.5}} \\

\blue{H\textsubscript{2}OT} w. MixSTE &[121, 81] &[0, 5] &33.78 & 129.64 (\textbf{\blue{$\downarrow$ 53.2\%}}) &82 (\textbf{\blue{$\uparrow$ 100.0\%}}) &5.7 (\textbf{\blue{$\downarrow$ 50.0\%}}) & \ \ 8.7 (\textbf{\blue{$\downarrow$ 51.4\%}}) &40.8 \\

\blue{H\textsubscript{2}OT} w. MixSTE &[121, 81] &[0, 7] &33.78 & 141.05 (\textbf{\blue{$\downarrow$ 49.1\%}}) &80 (\textbf{\blue{$\uparrow$ 95.1\%}}) &6.4 (\textbf{\blue{$\downarrow$ 43.9\%}}) & \ \ 9.5 (\textbf{\blue{$\downarrow$ 46.9\%}}) &40.6 \\

\blue{H\textsubscript{2}OT} w. MixSTE &[121, 81] &[3, 5] &33.78 & 173.14 (\textbf{\blue{$\downarrow$ 37.6\%}}) &64 (\textbf{\blue{$\uparrow$ 56.1\%}}) &7.8 (\textbf{\blue{$\downarrow$ 31.6\%}}) &11.3 (\textbf{\blue{$\downarrow$ 36.9\%}}) &40.9 \\

\blue{H\textsubscript{2}OT} w. MixSTE &[121, 81] &[3, 7] &33.78 & 184.55 (\textbf{\blue{$\downarrow$ 33.4\%}}) &62 (\textbf{\blue{$\uparrow$ 51.2\%}}) &8.2 (\textbf{\blue{$\downarrow$ 28.1\%}}) &12.0 (\textbf{\blue{$\downarrow$ 33.0\%}}) &40.8 \\

\midrule [0.5pt]
\midrule [0.5pt]

\blue{H\textsubscript{2}OT} w. MixSTE &[189, 121, 81] &[0, 3, 5] &33.78 & 153.89 (\textbf{\blue{$\downarrow$ 44.5\%}}) &72 (\textbf{\blue{$\uparrow$ 75.6\%}}) &6.9 (\textbf{\blue{$\downarrow$ 39.5\%}}) &10.8 (\textbf{\blue{$\downarrow$ 39.7\%}}) &40.9 \\

\blue{H\textsubscript{2}OT} w. MixSTE &[121, 81, 49] &[0, 3, 5] &33.78 & 104.54 (\textbf{\blue{$\downarrow$ 62.3\%}}) &81 (\textbf{\blue{$\uparrow$ 97.6\%}}) &5.1 (\textbf{\blue{$\downarrow$ 55.3\%}}) & \ \ 7.2 (\textbf{\blue{$\downarrow$ 59.8\%}}) &40.7 \\

\blue{H\textsubscript{2}OT} w. MixSTE &[121, 81, 27] &[0, 3, 5] &33.78 & \ \ 95.13 (\textbf{\blue{$\downarrow$ 65.7\%}}) &84 (\textbf{\blue{$\uparrow$ 104.9\%}}) &4.9 (\textbf{\blue{$\downarrow$ 57.0\%}}) & \ \ 6.7 (\textbf{\blue{$\downarrow$ 62.6\%}}) &41.6 \\

\midrule [0.5pt]

\blue{H\textsubscript{2}OT} w. MixSTE &[121, 81, 49] &[0, 3, 5] &33.78 & 104.54 (\textbf{\blue{$\downarrow$ 62.3\%}}) &84 (\textbf{\blue{$\uparrow$ 104.9\%}}) &5.1 (\textbf{\blue{$\downarrow$ 55.3\%}}) & \ \ 7.2 (\textbf{\blue{$\downarrow$ 59.8\%}}) &40.7 \\

\blue{H\textsubscript{2}OT} w. MixSTE &[121, 81, 49] &[0, 3, 7] &33.78 & 113.67 (\textbf{\blue{$\downarrow$ 59.0\%}}) &83 (\textbf{\blue{$\uparrow$ 102.4\%}}) &5.4 (\textbf{\blue{$\downarrow$ 52.6\%}}) & \ \ 7.8 (\textbf{\blue{$\downarrow$ 56.4\%}}) &40.9 \\

\blue{H\textsubscript{2}OT} w. MixSTE &[121, 81, 49] &[0, 5, 7] &33.78 & 125.08 (\textbf{\blue{$\downarrow$ 54.9\%}}) &71 (\textbf{\blue{$\uparrow$ 73.2\%}}) &5.8 (\textbf{\blue{$\downarrow$ 49.1\%}}) & \ \ 9.0 (\textbf{\blue{$\downarrow$ 49.7\%}}) &40.8 \\

\blue{H\textsubscript{2}OT} w. MixSTE &[121, 81, 49] &[3, 5, 7] &33.78 & 168.57 (\textbf{\blue{$\downarrow$ 39.2\%}}) &66 (\textbf{\blue{$\uparrow$ 61.0\%}}) &7.6 (\textbf{\blue{$\downarrow$ 33.3\%}}) &11.1 (\textbf{\blue{$\downarrow$ 38.0\%}}) &40.7 \\

\midrule [0.5pt]
\midrule [0.5pt]

\blue{H\textsubscript{2}OT} w. MixSTE &[189, 121, 81, 49] &[0, 3, 5, 7] &33.78 &149.32 (\textbf{\blue{$\downarrow$ 46.1\%}}) &74 (\textbf{\blue{$\uparrow$ 80.5\%}}) &6.7 (\textbf{\blue{$\downarrow$ 41.2\%}}) &10.2 (\textbf{\blue{$\downarrow$ 43.0\%}}) &40.8 \\

\blue{H\textsubscript{2}OT} w. MixSTE &[121, 81, 49, 27] &[0, 3, 5, 7] &33.78 &101.40 (\textbf{\blue{$\downarrow$ 63.4\%}}) &84 (\textbf{\blue{$\uparrow$ 104.9\%}}) &5.1 (\textbf{\blue{$\downarrow$ 55.3\%}}) & \ \ 7.7 (\textbf{\blue{$\downarrow$ 57.0\%}}) &41.3 \\

\bottomrule [1pt]

\end{tabular}
\label{table:parameters}
\end{table*}

\noindent \textbf{Inference Pipelines.}
In Table~\ref{table:pipeline}, we compare the efficiency and accuracy of different inference pipelines (mentioned in Sec~\ref{sec:VPTs}). 
We conduct experiments on MixSTE \cite{mixste} and MotionBERT \cite{motionbert} because both are designed for \textit{seq2seq} pipeline and can be evaluated on both \textit{seq2frame} and \textit{seq2seq} pipelines. 
As shown in the table, the \textit{seq2frame} can achieve better estimation accuracy by taking advantage of past and future information but lower efficiency due to repeated computation, \textit{i.e.}, 40.7mm vs. 40.9mm and 41 FPS vs. 9982 FPS for MixSTE (about 243$\times$ lower). 
Moreover, our method can reduce the computational costs and improve the inference speed of these two pipelines, while maintaining or obtaining better performance. 

For the \textit{seq2seq}, our method can reduce the FLOPs of MixSTE and MotionBERT by 57.4\% and 63.4\% and improve the FPS by 87.8\% and 47.6\%, respectively. 
Meanwhile, our method can improve performance by 0.5mm for MixSTE and only decrease it by 0.1mm (0.25\%) for MotionBERT. 
For the \textit{seq2frame}, our TPM w. MixSTE can reduce the FLOPs by 57.4\% and improve the FPS by 68.3\%, while bringing 0.3mm improvement. 
Additionally, our TPM w. MotionBERT can reduce 63.4\% FLOPs and improve 46.2\% FPS, while the estimation errors are reduced from 39.5mm to 39.3mm. 
Note that our method with TPM outperforms the one utilizing H\textsubscript{2}OT, largely because TRM in H\textsubscript{2}OT is a reverse operation that uses inadequate information to recover the full-length tokens. 

We also compare our method with our conference version \cite{hot}. 
As shown in Table~\ref{table:pipeline}, compared with \cite{hot}, our method can further reduce the FLOPs, GPU memory cost, and training time, while improving the FPS and maintaining the performance. 
For example, our H\textsubscript{2}OT w. MixSTE outperforms HoT w. MixSTE \cite{hot} by reducing FLOPs by 29.4\% (118.24G vs. 167.52G), GPU memory cost by 26.3\% (5.6G vs. 7.6G), and training time by 25.0\% (8.4 min/epoch vs. 11.2 min/epoch), while improving the FPS by 31.1\% (18745 vs. 14297) and MPJPE by 0.5mm (from 41.0mm to 40.5mm).
This demonstrates the effectiveness of our H\textsubscript{2}OT in reducing the computational costs, enabling stronger and faster VPT models on resource-limited devices. 

In the following ablations, we conduct experiments on the \textit{seq2seq} pipeline since it is more efficient and widely used in practice, and we choose MixSTE \cite{mixste} as the baseline since it is the first \textit{seq2seq} transformer-based architecture, and MotionBERT \cite{motionbert} and MotionAGFormer \cite{motionagformer} are its follow-ups. 

\noindent \textbf{Hyperparameters ($r$ and $b$).}
The number of representative tokens ($r$) and the block index of representative tokens ($b$) can be flexibly adjusted, thereby achieving the trade-off between computational costs and accuracy. 
Table~\ref{table:parameters} shows ablation studies on $r$ and $b$ in a hierarchical design under \textit{seq2seq} pipeline. 
We can see that decreasing $b$ and $r$ can reduce the FLOPs, and the best accuracy is achieved while $r = [121, 81]$ and $b = [0, 3]$. 
This indicates that appropriate $b$ and $f$ can bring a good trade-off between retaining important information and reducing redundant information for both the pruning and recovering stages. 
Therefore, the optimal hyper-parameters for our H\textsubscript{2}OT w. MixSTE are $r = [121, 81]$ and $b = [0, 3]$. 
We can also flexibly adjust $r$ and $b$ to achieve a speed-accuracy trade-off according to specific requirements. 

We also compare our method with MixSTE \cite{mixste} using the same number of representative tokens and approximately the same number of FLOPs. 
To achieve this, we set the input frame number of the original MixSTE to $F  =  81, 105, 121$. 
The results in Table~\ref{table:comparison} show that our method obtains better results, further demonstrating the importance of large receptive fields and the effectiveness of our method. 
Additionally, we compare our method with our conference version HoT \cite{hot} using similar hyperparameters and approximately the same number of FLOPs. 
Table~\ref{table:comparison} shows that our H\textsubscript{2}OT w. MixSTE using fewer parameters and FLOPs achieves better performance than HoT w. MixSTE \cite{hot}. 
This further demonstrates the effectiveness of our hierarchical design and efficient TPS and TRI. 

\begin{figure*}[tb!]
\centering
\includegraphics[width=1.0\linewidth]{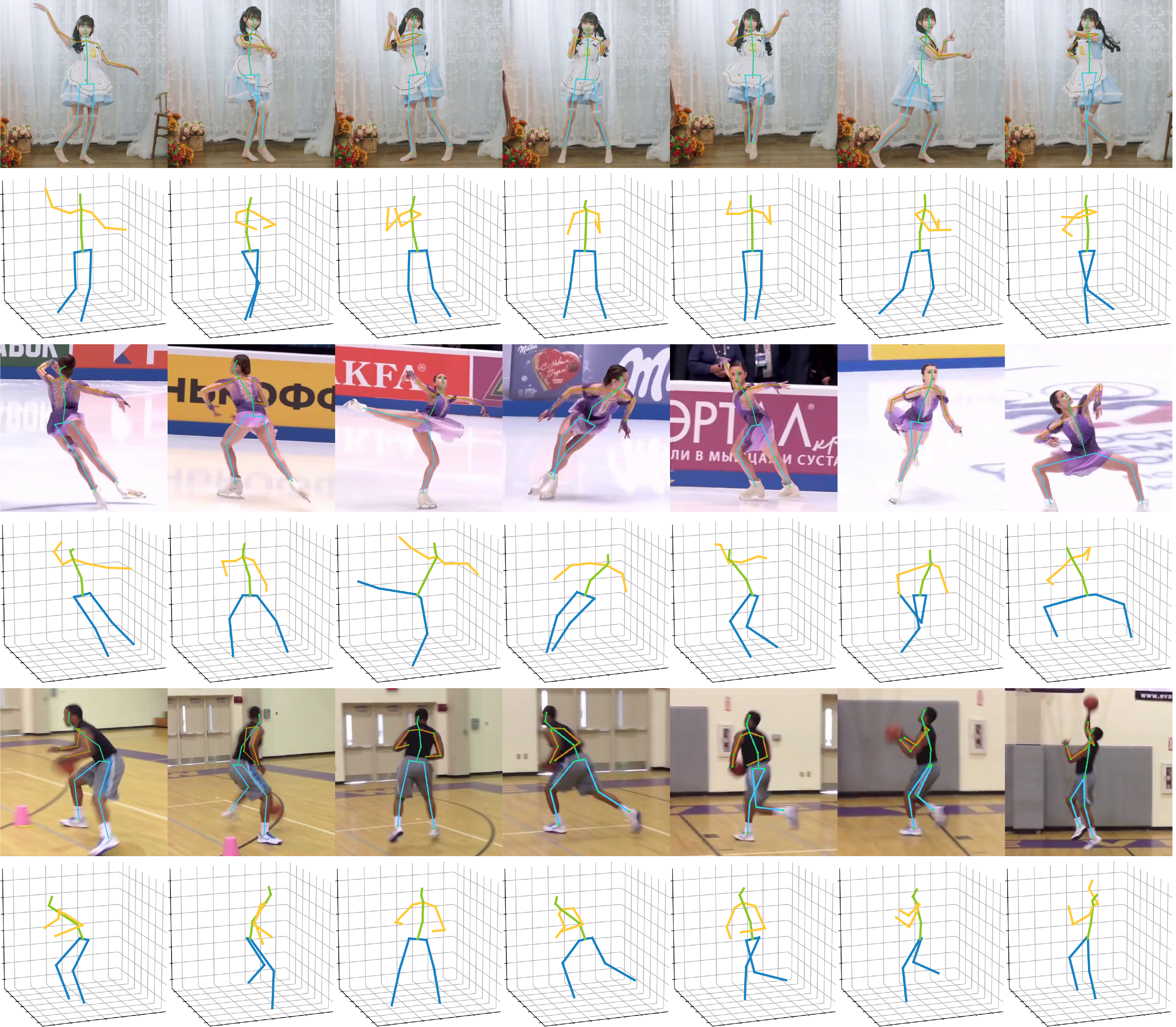}
\caption
{
    Qualitative results of our method on challenging in-the-wild videos. 
}
\label{fig:wild_supp}
\end{figure*}

\begin{figure*}[tb!]
\centering
\includegraphics[width=0.93\linewidth]{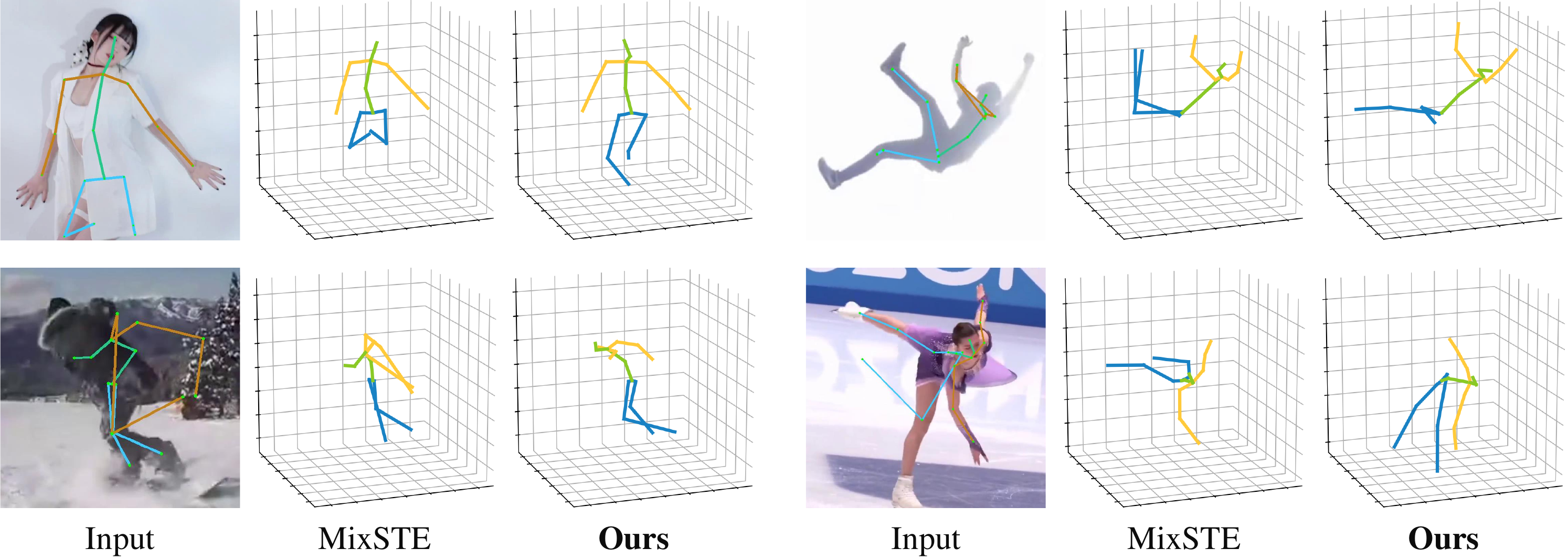}
\caption
{
    Failure cases in challenging scenarios. 
}
\label{fig:fail_supp}
\end{figure*}

\begin{figure*}[tb!]
\centering
\includegraphics[width=1.0\linewidth]{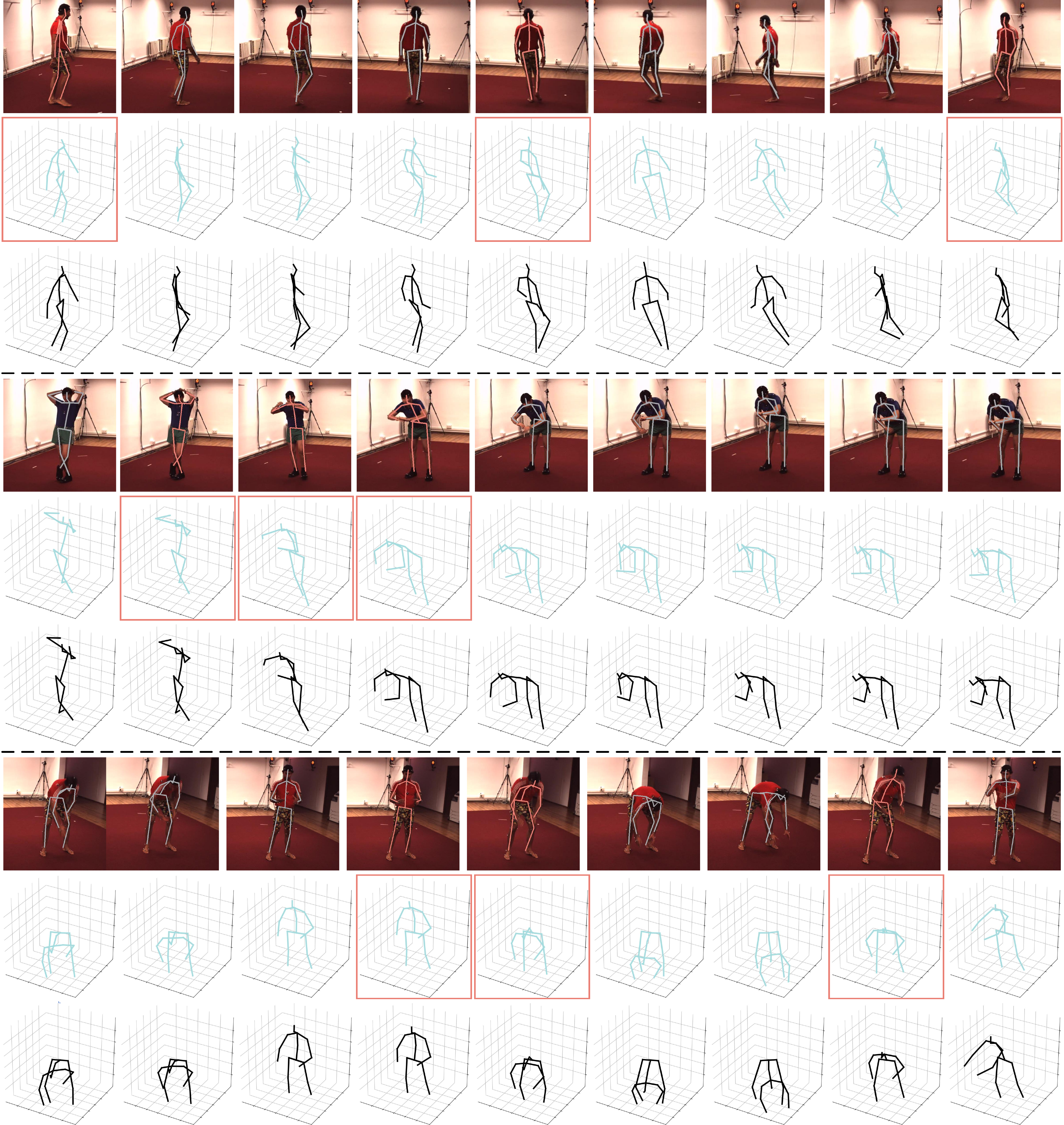}
\caption
{
    Visualization of input images, estimated 3D poses (cyan), and ground truth 3D poses (black) from three video sequences. 
    The 2D poses of selected frames are colored in red, and the 2D poses of pruned frames are colored in gray. 
    The 3D poses of selected frames are highlighted with red rectangular boxes. 
}
\label{fig:recovey_supp}
\end{figure*}

\noindent \textbf{Number of Recovered Tokens.}
In Table~\ref{table:recovered}, we conduct the ablation study on the number of recovered tokens ($f$) under \textit{seq2frame} pipeline. 
Since $f$ differs from the input frames, we evaluate the performance under the \textit{seq2frame} pipeline, which selects the 3D pose of the center frame as the final estimation. 
The results show that the performance remains almost unchanged when adjusting $f$. 
Therefore, we choose $f  =  243$, which is more efficient and can be evaluated under the \textit{seq2seq} pipeline. 

\begin{table}[t]
\footnotesize
\centering
\caption
{ 
    Comparison with MixSTE \cite{mixste} and HoT \cite{hot}. 
}
\setlength{\tabcolsep}{0.62mm}
\begin{tabular}{l|cllcc|c}
\toprule [1pt]
Method &$F$ &$b$ &$r$ &Param (M) &FLOPs (G) &MPJPE $\downarrow$ \\
\midrule [0.5pt]

MixSTE \cite{mixste} &81 &81 &- &33.70 &92.42 &42.7 \\

MixSTE \cite{mixste} &121 &121 &- &33.72 &138.05 &42.0 \\

MixSTE \cite{mixste} &105 &105 &- &33.71 &119.80 &42.3 \\

\midrule [0.5pt]

HoT w. MixSTE &243 &81 &0 &35.00 &109.76 &41.4 \\

HoT w. MixSTE &243 &81 &3 &35.00 &167.52 &41.0 \\

HoT w. MixSTE &243 &121 &0 &35.02 &153.26 &41.3 \\

HoT w. MixSTE &243 &121 &3 &35.02 &196.75 &41.1 \\

HoT w. MixSTE &243 &81 &1 &35.00 &121.31 &41.2 \\

HoT w. MixSTE &243 &89 &0 &35.00 &118.46 &41.7 \\
HoT w. MixSTE &243 &15 &3 &34.97 &119.28 &42.9 \\

\midrule [0.5pt]

\blue{H\textsubscript{2}OT} w. MixSTE &243 &[121, 81] &[0, 3] &33.78 &118.23 &\textbf{\red{40.5}} \\

\bottomrule [1pt]
\end{tabular}
\label{table:comparison}
\end{table}

\begin{table}[t]
\footnotesize
\centering
\caption
{ 
Ablation study on the number of recovered tokens ($f$) under \textit{seq2frame} pipeline.
}  
\setlength{\tabcolsep}{2.10mm}
\begin{tabular}{l|cc|c}
\toprule [1pt]
Method &Param (M) &FLOPs (G) &MPJPE $\downarrow$ \\
\midrule [0.5pt]

MixSTE \cite{mixste} &33.78 &277.25 &40.9 \\

\midrule [0.5pt]

\blue{H\textsubscript{2}OT} w. MixSTE ($f = 9$) &33.78 &118.23 &41.2 \\

\blue{H\textsubscript{2}OT} w. MixSTE ($f = 27$) &33.78 &118.23 &40.7 \\

\blue{H\textsubscript{2}OT} w. MixSTE ($f = 81$) &33.78 &118.23 &40.7 \\

\blue{H\textsubscript{2}OT} w. MixSTE ($f = 243$) &33.78 &118.23 &\textbf{\red{40.4}} \\

\bottomrule [1pt]
\end{tabular}
\label{table:recovered}
\end{table}

\begin{table}[t]
\scriptsize
\centering
\caption
{ 
Comparison of parameters (M), FLOPs (G), and MPJPE with SOTA VPTs on Human3.6M. 
Here, $F$ denotes the number of input frames. 
$^{*}$ indicates our re-implementation.
}
\setlength{\tabcolsep}{0.40mm} 
\begin{tabular}{l|cllc}
\toprule [1pt]
Method &$F$ &Param &FLOPs &MPJPE $\downarrow$  \\
\midrule [0.5pt]
PoseFormer (ICCV'21) \cite{poseformer} &81 &9.60 &1.63 &44.3 \\

Strided (TMM'22) \cite{stride} &351 &4.35 &1.60 &43.7 \\

P-STMO (ECCV'22) \cite{pstmo} &243 &7.01 &1.74 &42.8 \\

Einfalt \emph{et al.} (WACV'23) \cite{einfalt2023uplift} &351 &10.36 &1.00 &44.2 \\

HDFormer (IJCAI'23) \cite{chen2023hdformer} &96 &3.70 &1.18 &42.6 \\

STCFormer (CVPR'23) \cite{tang20233d} &243 &18.93 &156.22 &40.5 \\

KTPFormer (CVPR'24) \cite{ktpformer} &243 &33.65 &277.27 &40.1 \\

\midrule [0.5pt]

MHFormer (CVPR'22) \cite{mhformer} &351 &31.52 &14.15 &43.0 \\

TPC w. MHFormer (CVPR'24)
\cite{hot} &351 &31.52 &\ \ 8.22 (\textbf{\blue{$\downarrow$ 41.9\%}}) &{43.0} \\

\blue{TPM} w. MHFormer (\textbf{Ours}) &351 &31.52 &\ \ 6.74 (\textbf{\blue{$\downarrow$ 52.4\%}}) &43.2 \\

\midrule [0.5pt]

MixSTE (CVPR'22) \cite{mixste} &243 &33.78 &277.25 &40.9 \\

HoT w. MixSTE (CVPR'24) \cite{hot} &243 &35.00 &167.52 (\textbf{\blue{$\downarrow$ 39.6\%}}) &{41.0} \\

\blue{H\textsubscript{2}OT} w. MixSTE (\textbf{Ours}) &243 &33.78 &118.23 (\textbf{\blue{$\downarrow$ 57.4\%}}) &40.5 \\

\midrule [0.5pt]

MotionBERT (ICCV'23) \cite{motionbert} &243 &16.00 &131.09 &39.2 \\

MotionBERT (ICCV'23) \cite{motionbert}$^{*}$ &243 &16.00 &131.09 &39.8 \\

HoT w. MotionBERT (CVPR'24) \cite{hot} & 243 &16.35 &\ \ 63.21 (\textbf{\blue{$\downarrow$ 51.8\%}}) &39.8 \\

\blue{H\textsubscript{2}OT} w. MotionBERT (\textbf{Ours}) &243 &16.00 &\ \ 47.99 (\textbf{\blue{$\downarrow$ 63.4\%}}) &39.9 \\

\midrule [0.5pt]

MotionAGFormer (WACV'24) \cite{motionagformer} &243 &11.72 &\ \ 95.99 &38.4 \\

HoT w. MotionAGFormer (CVPR'24) \cite{hot} &243 &11.83 &\ \ 60.56 (\textbf{\blue{$\downarrow$ 36.9\%}}) &38.9 \\

\blue{H\textsubscript{2}OT} w. MotionAGFormer (\textbf{Ours}) &243 &11.72 &\ \ 38.87 (\textbf{\blue{$\downarrow$ 59.5\%}}) &38.5 \\

\bottomrule [1pt]
\end{tabular}
\label{table:h36m}
\end{table}

\subsection{Comparison with state-of-the-art methods}
Current SOTA performance on Human3.6M is achieved by transformer-based architectures. 
We compare our method with them by adding it to four very recent VPTs: 
MHFormer \cite{mhformer}, MixSTE \cite{mixste}, MotionBERT \cite{motionbert}, and MotionAGFormer \cite{motionagformer}. 
These four models significantly outperform previous works at the cost of high computational complexity, thus we choose them as baselines to evaluate our method. 
The comparisons are shown in Table~\ref{table:h36m}. 
For example, compared with MixSTE \cite{mixste}, our H\textsubscript{2}OT w. MixSTE obtains better performance with 0.5mm improvements while reducing computational costs by 57.4\% in FLOPs. 
We also compare our method with diffusion-based VPTs (D3DP \cite{shan2023diffusion} and DiffPose \cite{gong2023diffpose}) in Table \ref{table:diff}. 
It can be found that our H\textsubscript{2}OT with D3DP reduces FLOPs by 57.4\% and increases FPS by 40.7\%, with only a 0.1mm decrease in performance.
Additionally, H\textsubscript{2}OT with DiffPose achieves a 57.4\% reduction in FLOPs and a 44.5\% improvement in FPS with a 0.2mm decrease in performance. 
These experiments demonstrate the effectiveness and efficiency of our method, while revealing that existing VPTs incur redundant computational costs that contribute little to the estimation accuracy and even decrease the accuracy. 

Furthermore, we conduct experiments on Human3.6M at varying FPS (50Hz, 25Hz, 16.7Hz, and 12.5Hz), corresponding to downsampling rates of 1, 2, 3, and 4, respectively. 
The results are shown in Table \ref{table:fps}. 
We can see a performance improvement at lower FPS, which can be attributed to reduced temporal redundancy and an expanded temporal receptive field \cite{pstmo}. 
However, we observe that our H\textsubscript{2}OT w. MixSTE shows slightly inferior performance compared to the original MixSTE at low FPS. 
This can be attributed to the fact that in low-FPS scenarios, temporal information is sparse, leaving less redundant data that can be removed through token pruning. 
Despite this, our method maintains its core advantage: a remarkable 57.4\% reduction in computational cost while preserving competitive performance. 
This makes our method particularly valuable for real-world applications where computational efficiency is paramount. 

\begin{table}[t]
\scriptsize
\centering
\caption
{
Comparison of efficiency and accuracy with diffusion-based methods on Human3.6M. 
Frame per second (FPS) was computed on a single GeForce RTX 3090 GPU. 
}
\setlength{\tabcolsep}{0.29mm}
\begin{tabular}{l|cll|c}
\toprule [1pt]
Method &Param (M) &FLOPs (G) &FPS &MPJPE $\downarrow$  \\

\midrule [0.5pt]

D3DP \cite{shan2023diffusion} ($H=5$, $K=5$) &34.84 &277.26 &820 &39.7 \\
\blue{H\textsubscript{2}OT} w. D3DP &34.84 &118.25 (\textbf{\blue{$\downarrow$ 57.4\%}}) &1154 (\textbf{\blue{$\uparrow$ 40.7\%}})  &{39.8} \\

\midrule [0.5pt]

DiffPose \cite{gong2023diffpose} &33.78 &277.25 &6522 &39.5 \\

\blue{H\textsubscript{2}OT} w. DiffPose &33.78  & 118.23 (\textbf{\blue{$\downarrow$ 57.4\%}}) &9426 (\textbf{\blue{$\uparrow$ 44.5\%}})  &{39.7} \\

\bottomrule [1pt]

\end{tabular}
\label{table:diff}
\end{table}

\begin{table}[t]
\footnotesize
\centering
\caption
{ 
Results on Human3.6M at different FPS settings. 
}
\setlength{\tabcolsep}{4.70mm}
\begin{tabular}{l|cccc}
\toprule [1pt]
FPS &50.0 &25.0 &16.7 &12.5  \\

\midrule [0.5pt]

MixSTE &40.9 &39.6 &39.3 &39.2  \\

\blue{H\textsubscript{2}OT} w. MixSTE &40.5 &40.2 &40.0 &40.8 \\

\bottomrule [1pt]

\end{tabular}
\label{table:fps}
\end{table}

\begin{table}[t]
\scriptsize
\centering
\caption
{
Quantitative comparison with SOTA VPTs on MPI-INF-3DHP. 
}
\setlength{\tabcolsep}{1.95mm} 
\begin{tabular}{@{}lc|ccc@{}}
\toprule [1pt]
Method &$F$ & PCK $\uparrow$  & AUC $\uparrow$  & MPJPE $\downarrow$  \\
\midrule [0.5pt]

PoseFormer (ICCV'21) \cite{poseformer} &9 &{88.6} &{56.4} &{77.1} \\ 

P-STMO (ECCV'22) \cite{pstmo} &81 &{97.9} &{75.8} &{32.2} \\

Einfalt \emph{et al.} (WACV'23) \cite{einfalt2023uplift} &81 &95.4 &67.6 &46.9 \\

HDFormer (IJCAI'23) \cite{chen2023hdformer} &81 &98.7 &72.9 &37.2 \\

STCFormer (CVPR'23) \cite{tang20233d} &81 &98.7 &83.9 &23.1 \\

KTPFormer (CVPR'24) \cite{ktpformer} &81 &98.9 &85.9 &16.7 \\

\midrule [0.5pt]

MHFormer (CVPR'22) \cite{mhformer} &9 &{93.8} &{63.3} &{58.0} \\

TPC w. MHFormer (CVPR'24) \cite{hot} &9 &94.0 &63.3 &58.4 \\

\blue{TPM} w. MHFormer (\textbf{Ours}) &9 &94.1 &63.6 &57.8 \\

\midrule [0.5pt]

MixSTE (CVPR'22) \cite{mixste} &27 &{94.4} &{66.5} &{54.9} \\

HoT w. MixSTE (CVPR'24) \cite{hot} &27 &{94.8} &{66.5} &53.2 \\

\blue{H\textsubscript{2}OT} w. MixSTE (\textbf{Ours}) &27 &95.2 &66.4 &53.0 \\

\midrule [0.5pt]

MotionAGFormer (WACV'24) \cite{motionagformer} &81 &98.3 &84.2 &18.2 \\

HoT w. MotionAGFormer (CVPR'24) \cite{hot} &81 &99.0 &84.5 &18.8 \\

\blue{H\textsubscript{2}OT} w. MotionAGFormer (\textbf{Ours}) &81 &99.1 &85.2 &18.0 \\

\bottomrule [1pt]
\end{tabular}
\label{table:3dhp}
\end{table}

We further evaluate our method on MPI-INF-3DHP dataset in Table~\ref{table:3dhp}. 
For a fair comparison, following \cite{mhformer,mixste,motionagformer}, we implement our method on them with the same number of input frames: MHFormer with \{$F = 9$, $r = [5, 3]$, $b = [0, 1]$\},  MixSTE with \{$F = 27$, $r = [13, 9]$, $b = [0, 3]$\}, and MotionAGFormer with \{$F = 81$, $r = [41, 27]$, $b = [0, 7]$\}. 
It can be found that our method (TPM w. MHFormer, H\textsubscript{2}OT w. MixSTE, and H\textsubscript{2}OT w. MotionAGFormer) achieves competitive performance compared with our original version \cite{hot} and the original VPTs \cite{mhformer,mixste,motionagformer}. 
These results demonstrate the effectiveness of our method in both indoor and outdoor scenes, and our method can also work well with a small temporal receptive field. 

\subsection{Qualitative Results}
Figure~\ref{fig:wild_supp} shows the qualitative results on challenging in-the-wild videos. 
These results confirm the ability of our method to produce accurate 3D pose estimations. 
However, in challenging scenarios, there are some failure cases where our method cannot accurately estimate 3D human poses due to factors such as partial body visibility, rare poses, and significant errors in the 2D detector (Figure~\ref{fig:fail_supp}). 
We also provide visualizations of recovering 3D human poses in Figure~\ref{fig:recovey_supp}, which illustrate that our method can predict realistic 3D human poses of the entire sequence. 

\section{Conclusion}
This paper presents Hierarchical Hourglass Tokenizer (H\textsubscript{2}OT), a hierarchical plug-and-play \textit{pruning-and-recovering} framework for efficient transformer-based 3D human pose estimation from videos. 
Our method reveals that maintaining the full pose sequence is unnecessary, and using a few pose tokens of representative frames can achieve both high efficiency and estimation accuracy. 
Comprehensive experiments demonstrate that our method is compatible and general. 
It can be easily incorporated into common VPT models on both \textit{seq2seq} and \textit{seq2frame} pipelines while effectively accommodating various token pruning and recovery strategies, thereby highlighting its potential for using future ones. 
We hope H\textsubscript{2}OT can enable the creation of stronger and faster VPTs. 

\section*{Acknowledgments}

This work was supported by the National Natural Science Foundation of China (No. 62373009), the Shenzhen Innovation in Science and Technology Foundation for The Excellent Youth Scholars (No. RCYX20231211090248064), the Agency for Science, Technology and Research (A*STAR) with a project number R24I6IR138, the EU Horizon projects ELIAS (No. 101120237) and ELLIOT (No. 101214398), and by the MUR PNRR project FAIR (PE00000013) funded by the NextGenerationEU. 

\bibliographystyle{IEEEtran}
\bibliography{ref.bib}

\begin{IEEEbiography}[{\includegraphics[width=1in,height=1.25in,clip,keepaspectratio]{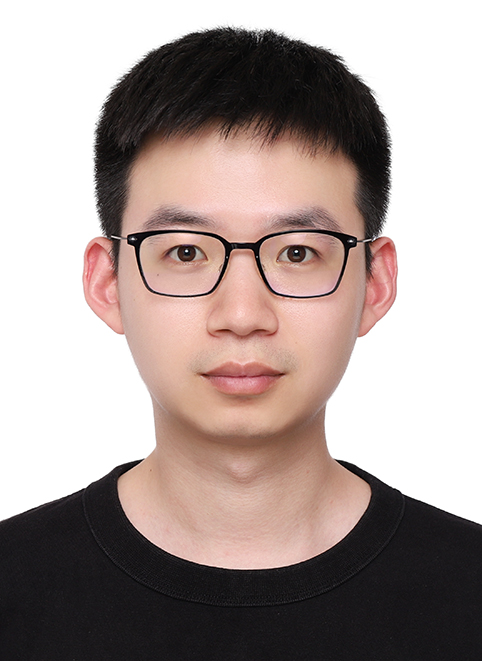}}]{Wenhao Li} is currently a Postdoctoral Researcher at the College of Computing and Data Science, Nanyang Technological University, Singapore.  
He received the Ph. D. degree from the School of Computer Science, Peking University, China. 
His research interests lie in deep learning, machine learning, and their applications to computer vision.
\end{IEEEbiography}

\begin{IEEEbiography}[{\includegraphics[width=1in,height=1.25in,clip,keepaspectratio]{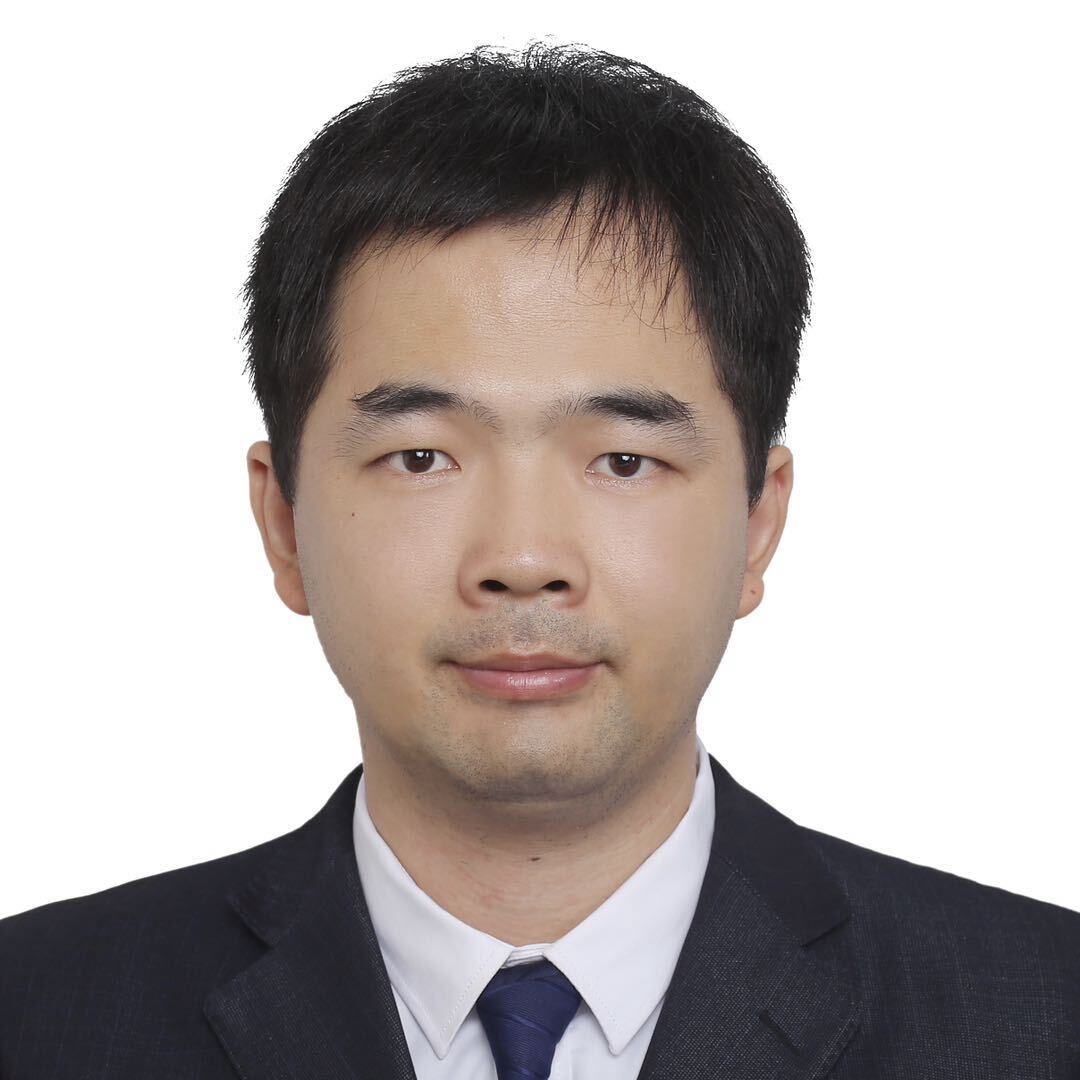}}]{Mengyuan Liu} is an Assistant Professor at Peking University. He received a Ph.D. degree in 2017 under the supervision of Prof. H. Liu from the School of EE\&CS, Peking University (PKU), China. His research interests include human action recognition, human motion prediction, and human motion generation using RGB, depth, and skeleton data. Related methods have been published in T-IP, T-CSVT, T-MM, PR, CVPR, ECCV, ACM MM, AAAI, and IJCAI. He has been invited to be a Technical Program Committee (TPC) member for ACPR, ACM MM, and AAAI. 
\end{IEEEbiography}

\begin{IEEEbiography}[{\includegraphics[width=1in,height=1.25in,clip,keepaspectratio]{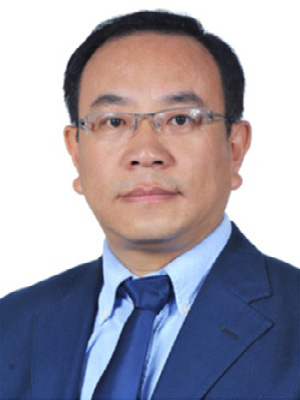}}]{Hong Liu} is a full professor at Peking University, and serves as Co-director of the Center of Embodied Intelligence and Robotics, Institute of AI, Peking University and is also the founding dean of Chongqing Liangjiang School of Artificial Intelligence. He serves as the vice president of the Chinese Association for Artificial Intelligence (CAAI), and the founding Editor-in-Chief for the top international journal of CAAI Transactions on Intelligence Technology.  He earned the National Aerospace Science and Technology Progress Award, Geneva International Invention Award, the First Price in Nature Science of Wu Wenjun Artificial Intelligence Science and Technology Award, the First Price in Natural Science of Shenzhen Science and Technology Award. Hong Liu has engaged in the research of artificial intelligence, robotics, machine Learning and intelligent human robot interaction for more than twenty years. He published over 300 papers in international journals and proceedings in the above areas and these papers have been cited over 16000 times. 
\end{IEEEbiography}

\begin{IEEEbiography}[{\includegraphics[width=1in,height=1.25in,clip,keepaspectratio]{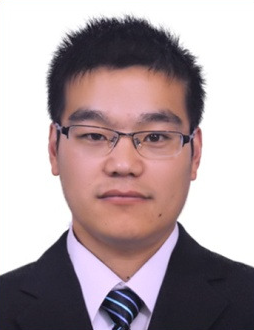}}]{Pichao Wang} received the Ph.D. degree in computer science from the University of Wollongong, Australia. He is currently a Senior Applied Scientist in Amazon AGI, USA. He has authored 120+ peer reviewed papers, including those in highly regarded journals and conferences such as IEEE TPAMI, IJCV, CVPR, ICCV, ECCV, ICLR, NeurIPS, AAAI, ACM MM, etc. He is the recipient of CVPR2022 Best Student Paper Award. He was named AI 2000 Most Influential Scholar during 2012-2022 by Miner, due to his contributions in the field of multimedia. He is also in the list of World’s Top 1\% Scientists named by Stanford University.  
\end{IEEEbiography}

\begin{IEEEbiography}[{\includegraphics[width=1in,height=1.25in,clip,keepaspectratio]{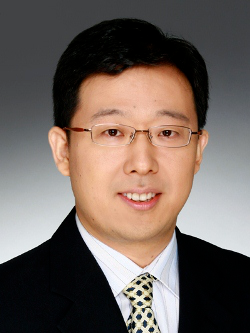}}]{Shijian Lu} is an Assistant Professor with the College of Computing and Data Science at Nanyang Technological University, Singapore. He received his PhD in electrical and computer engineering from the National University of Singapore. His major research interests include image and video analytics, visual intelligence, and machine learning.
\end{IEEEbiography}

\begin{IEEEbiography}[{\includegraphics[width=1in,height=1.25in,clip,keepaspectratio]{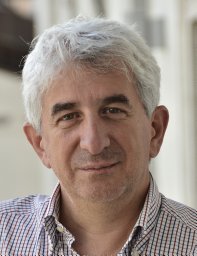}}]{Nicu Sebe} is a Professor in the University of Trento, Italy, where he is leading the research in the areas of multimedia analysis and human behavior understanding. He was the General Co-Chair of the IEEE FG 2008 and ACM Multimedia 2013 and 2022. He was a program chair of ACM Multimedia 2011 and 2007, ECCV 2016, ICCV 2017 and ICPR 2020. He is a fellow of IAPR and of ELLIS. 
\end{IEEEbiography}

\end{document}